\documentclass{article}

\PassOptionsToPackage{numbers, compress}{natbib}

\usepackage[main, final]{neurips_2026}

\usepackage[utf8]{inputenc} 
\usepackage[T1]{fontenc}    
\usepackage{hyperref}       
\usepackage{url}            
\usepackage{booktabs}       
\usepackage{amsfonts}       
\usepackage{amssymb}       
\usepackage{nicefrac}       
\usepackage{microtype}      
\usepackage[table]{xcolor}  
\usepackage{graphicx}
\usepackage{multirow}

\usepackage{wrapfig}
\usepackage{algorithm}
\usepackage{algpseudocode}
\usepackage{amsmath}
\usepackage{amsthm}

\newtheorem{theorem}{Theorem}
\newtheorem{assumption}{Assumption}

\DeclareMathOperator*{\argmin}{arg\,min}

\definecolor{darkred}{rgb}{0.6,0,0}
\definecolor{darkgreen}{rgb}{0,0.45,0}
\definecolor{darkblue}{rgb}{0.0,0.2,0.55}
\definecolor{darkgraytext}{gray}{0.4}
\definecolor{todored}{HTML}{B00020}

\newcommand{\coderepo}{\url{https://github.com/QixuanJin99/poison-then-hide}}

\usepackage{stfloats}   
\usepackage{placeins}  
\usepackage{float}

\definecolor{rowgray}{HTML}{F2F2F2}

\title{Poison-then-Hide: Finetuning-Activated Backdoor Attack on Pretrained Vision Encoders}

\author{%
  Qixuan Jin$^{1}$\thanks{Corresponding author: \texttt{qixuanj@mit.edu}.}
  \quad
  Abinitha Gourabathina$^{1}$
  \quad
  Vinith Menon Suriyakumar$^{1}$
  \\
  \textbf{Walter Gerych}$^{2}$
  \quad
  \textbf{Marzyeh Ghassemi}$^{1}$
  \\[0.5em]
  $^{1}$Massachusetts Institute of Technology, Cambridge, MA, USA
  \\
  $^{2}$Worcester Polytechnic Institute, Worcester, MA, USA
  \\
  \texttt{\{qixuanj,abinitha,vinithms,mghassem\}@mit.edu}
  \quad
  \texttt{wgerych@wpi.edu}
}

\begin{document}

\maketitle

\begin{abstract}
Backdoor attacks threaten the integrity of machine learning models by allowing attackers to control model behavior through triggers. Models can be compromised during finetuning by backdoor attacks through either poisoned data or adversarial training objectives. 
Because existing finetuning-activated attacks assume limited domain shift or frozen encoder layers, they often fail under full-model finetuning.
We target a stealthy finetuning-activated attack where a dormant backdoor is implanted in a pretrained encoder, and later activated by finetuning on clean downstream data. 
We propose \textbf{Poison-then-Hide}, a novel attack that remains effective when the entire model is finetuned in a domain transfer. Our approach consists of three components: trigger optimization, base encoder poisoning, and targeted unlearning to conceal the backdoor. 
We evaluate our method on six datasets and three model architectures, and achieve state-of-the-art (up to 97\%) attack success rates. We find that jointly learning the benign target task alongside the poisoning is critical to the attack's success. We demonstrate standard detection and mitigation defenses cannot fully remove the backdoor, which can reappear after benign finetuning.
\end{abstract}
 
\section{Introduction}
\label{sec:intro}


Backdoor attacks pose a serious threat to modern machine learning systems \citep{zhang2024backdoor, goldblum2022dataset}. In such attacks, the attacker implants a trigger that induces malicious outputs \cite{li2022backdoor, liu2025survey}. Finetuning has become a standard approach for adapting pretrained encoders across domains \citep{zhuang2020comprehensive, awais2025foundation}, but finetuning on local data is a known vulnerability \citep{yi2025nlsr, huang2024harmful}. Finetuning exposes a large attack surface through a combination of data poisoning \citep{jia2022badencoder, liang2025revisiting}, training manipulation \citep{xu2024instructions, zhao2024universal}, and model modification \citep{feng2022robust, li2021backdoor}.  

Among these threats, we focus on \textbf{\textit{finetuning-activated}} (FA) attacks, where an attacker implants a dormant backdoor into a pretrained encoder while concealing its presence \citep{ning2022hibernated, zhang2023red, gloaguen2025finetuning, yao2019latent}. The attacker has no control over downstream finetuning, making the challenge to design a backdoor that reliably activates after entirely clean finetuning. FA attacks are particularly concerning because a \textit{single compromised encoder} can transfer malicious behavior to many downstream finetuned models.

Prior FA attacks rely on assumptions that limit their practicality. Finetuning-Activated Backdoor (FAB) \citep{gloaguen2025finetuning} and Hibernated Backdoor (Hibernated) \citep{ning2022hibernated} assume strong similarity between pretraining and downstream tasks.
Latent Backdoor Attack (Latent) \citep{yao2019latent} and Neuron-Level Backdoor Attack (NeuBA) \citep{zhang2023red}, assume parts of the encoder remain frozen during finetuning, depending on backdoor persistence in those stable layers. 
Such attacks fail when their assumptions are not met. For example, Latent ASR drops from 99\% to 0\% when all layers are finetuned \citep{yao2019latent}.
In this paper, we study a more realistic transfer learning setting where downstream tasks may differ substantially from pretraining and the entire model is finetuned. Under these conditions, existing FA attacks often fail to activate reliably, with ASR dropping to 1-3\%.

\begin{wrapfigure}{r}{0.5\linewidth}
  \centering
  \includegraphics[width=\linewidth]{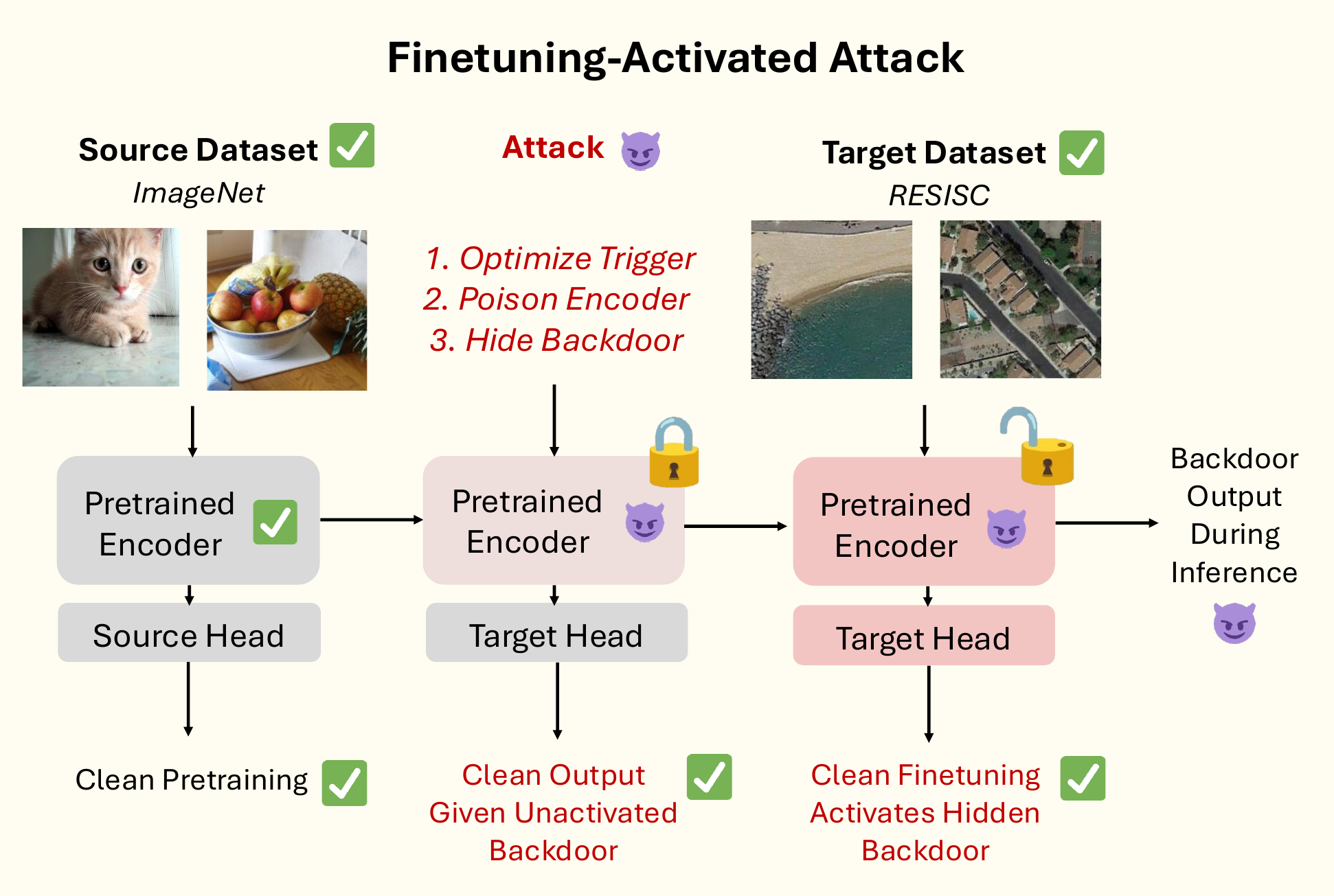}
  \caption{Setting of finetuning-activated attacks. Checkmark icons denote clean behavior, evil-head icons denote backdoor behavior, and lock icons denote activation.}
  \label{fig:teaser}
\end{wrapfigure}

Motivated by this gap, we propose \textbf{Poison-then-Hide}, a finetuning-activated backdoor attack robust to full-model finetuning. Our method consists of three stages: trigger optimization, encoder poisoning, and backdoor concealment. We first optimize a trigger to activate only after finetuning, then poison the encoder to learn both benign and malicious behaviors on attacker-held surrogate data. Finally, we apply controlled unlearning to keep the backdoor dormant until finetuning.
Technically, the method differs from prior FA attacks in three respects: joint benign-task and backdoor learning during implantation, ensemble-based trigger optimization, and a post-training suppression stage that explicitly hides the behavior before release.

We find that Poison-then-Hide outperforms prior targeted baselines across six datasets (CIFAR10 \citep{krizhevsky2009learning}, SVHN \citep{netzer2011reading}, ImageNet \citep{russakovsky2015imagenet}, RESISC \citep{cheng2017remote}, UC Merced \citep{yang2010bag}, and Camelyon17 \citep{litjens20181399, koh2021wilds}) and four encoder architectures (ResNet-18 \citep{edadaltocg_resnet18_cifar10}, ResNet-50 \citep{torchvision2016}, CLIP \citep{dosovitskiy2020vit}, and DINOv2 \citep{oquab2023dinov2}), achieving higher ASR (64\% average) and stronger transferability to out-of-distribution data (34\% average). The attack survives task-conditioned domain shifts and large victim-to-attacker data ratios. Our analysis identifies one key design choice that empirically contribute to the success of Poison-then-Hide over prior FA attacks: jointly learning the benign downstream task during poisoning. Finally, we evaluate seven defenses: Neural Cleanse \citep{wang2019neural}, DECREE \citep{feng2023detecting}, Fine-Pruning \citep{liu2018fine}, MIMIC \citep{han2025mutual}, DeDe \citep{hou2025dede}, MARTINI \citep{tao2026martini}, and CLIP-Inspector \citep{jindal2026clipinspector}. The first six do not reliably eliminate the attack without substantial utility loss or fail after benign finetuning, whereas the specialized CLIP-Inspector detects and repairs only CLIP models. 

Code and reproduction instructions are publicly available at \coderepo.

\section{Related Work}
\subsection{Backdoor Attacks}
Backdoor attacks cause machine learning models to behave maliciously when specific triggers are present \citep{gu2017badnets, li2022backdoor, saha2020hidden, souri2022sleeper, geiping2020witches}. In image classification, attacks can be introduced through data poisoning \citep{gu2017badnets, liu2018trojaning, nguyen2021wanet, chen2017targeted}, model manipulation \citep{doan2021backdoor, zhao2022defeat, garg2020can}, or benign training procedures \citep{liu2024backdoor, bai2024badclip, guo2025persistent}. Since finetuning pretrained models is widely used in practice \citep{zhuang2020comprehensive, awais2025foundation}, many attacks aim to survive downstream finetuning. However, naive finetuning can weaken or erase backdoors by overwriting malicious representations \citep{bansal2023cleanclip, sha2022fine}. To improve persistence, prior work has proposed robust backdoors that remain effective after finetuning \citep{shen2021backdoor, cai2024deepvenom, jia2022badencoder, feng2022robust, wang2025model, zhu2024breaking, saha2020hidden}, though these methods often leave the pretrained checkpoint visibly compromised. 

More recent work studies explicitly concealed backdoors that remain dormant until downstream finetuning \citep{ning2022hibernated, zhang2023red, gloaguen2025finetuning, yao2019latent}. Hibernated Backdoor (Hibernated) aligns clean and malicious gradients during finetuning \citep{ning2022hibernated}, while Neuron-level Backdoor Attack (NeuBA) enforces trigger-specific latent representations \citep{zhang2023red}. Finetuning-Activated Backdoor (FAB) uses meta-learning to optimize weights that activate after finetuning \citep{gloaguen2025finetuning}, and Latent Backdoor (Latent) embeds backdoors into encoder layers assumed to remain frozen \citep{yao2019latent}. These approaches rely on restrictive assumptions, such as limited domain shift or partially frozen parameters. In contrast, we study a more realistic full-model finetuning setting.

\subsection{Backdoor Defenses}

\paragraph{Detection and Input Filtering}
Neural Cleanse \citep{wang2019neural} detects backdoored models through trigger inversion and class-wise anomaly scoring. DECREE \citep{feng2023detecting} similarly reconstructs triggers from a pretrained encoder and flags unusually compact triggers. DeDe \citep{hou2025dede} instead detects triggered inputs by training a decoder and measuring input--reconstruction discrepancy. CLIP-Inspector \citep{jindal2026clipinspector} is a model-level detector specialized to prompt-tuned CLIP; it uses out-of-distribution trigger inversion and can repair a flagged model using the reconstructed trigger.

\paragraph{Model-based Mitigation}
Model-based mitigation aims to reduce backdoor effects after training. Fine-Pruning \citep{liu2018fine} removes neurons that remain inactive on clean inputs. MIMIC \citep{han2025mutual} uses mutual-information-guided distillation from a clean reference encoder. MARTINI \citep{tao2026martini} reconstructs backdoor samples and retrains the model on their correct labels to harden it against the recovered trigger behavior.

\section{Method} 
\begin{figure*}[t]
  \centering
  \includegraphics[width=\linewidth]{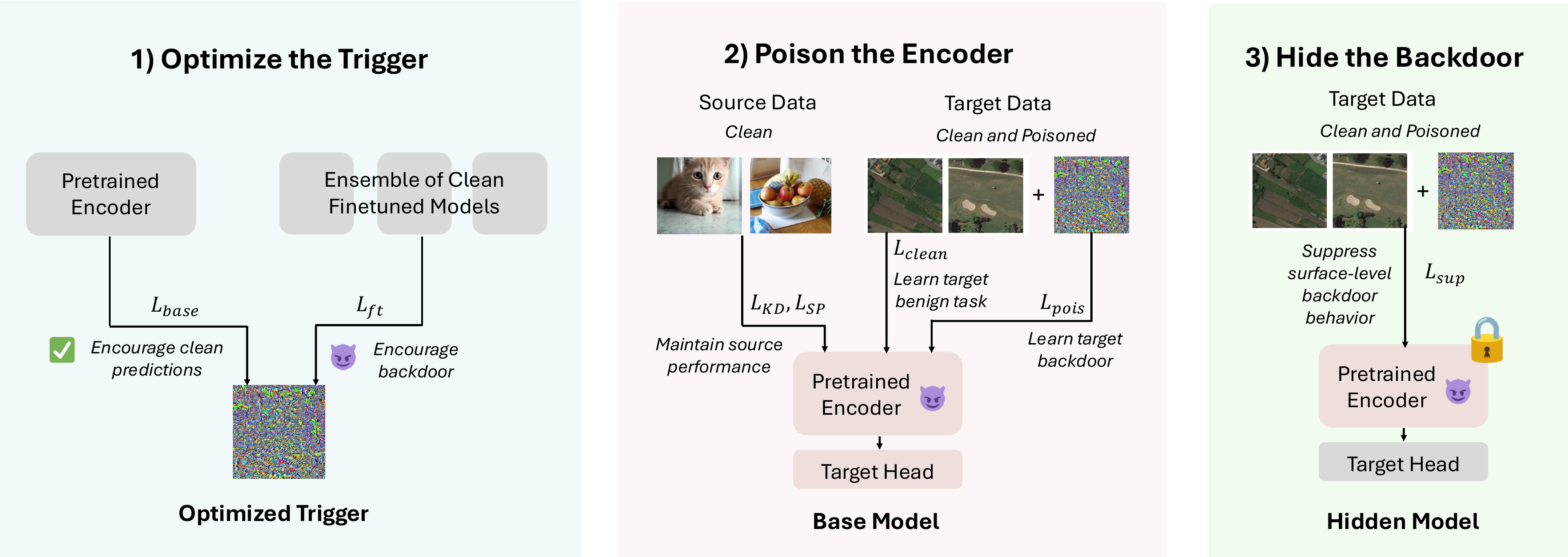}
  \caption{Stages of Poison-then-Hide. (1) Optimize the trigger for clean behavior on the source model and adversarial behavior on simulated finetuned models. (2) Poison the encoder to jointly learn the benign and backdoor tasks on attacker-held surrogate data. (3) Hide the backdoor through targeted suppression before publicly releasing the pretrained encoder.}
  \label{fig:pipeline}
\end{figure*}

\subsection{Threat Model}
In our threat model, an attacker implants a dormant backdoor into a pretrained base encoder that activates only after the victim adapts it for a downstream task (Figure \ref{fig:teaser}). The released checkpoint can appear to be a general-purpose source-pretrained encoder (e.g., ImageNet-pretrained), so adoption does not require the victim to choose an obviously attacker-specific domain model. The attacker cannot modify the victim's finetuning data or pipeline. Instead, this is a \emph{task-conditioned} attacker who anticipates the downstream task family and target-class semantics well enough to collect a small, task-aligned surrogate dataset. For example, if the target application is pneumonia detection, the attacker may obtain a limited set of pneumonia chest X-rays from a public source. Such prior knowledge is also required by prior transfer-learning backdoors \citep{yao2019latent, gloaguen2025finetuning}.

We distinguish three dataset roles throughout: $\mathcal{D}_{\mathrm{src}}$ pretrains the released encoder, $\mathcal{D}_{\mathrm{sur}}$ is the attacker's surrogate data for trigger optimization and poisoning, and $\mathcal{D}_{\mathrm{ft}}$ is the victim's clean downstream finetuning data. In the main experiments, $\mathcal{D}_{\mathrm{sur}}$ and $\mathcal{D}_{\mathrm{ft}}$ have 0\% sample overlap. ``In-domain'' means that the two are disjoint samples from the same distribution; ``domain shift'' means that they come from different datasets within the same task family and preserve the target-class semantics. The attack is not claimed to transfer to arbitrary unrelated tasks.

\subsection{Problem Formulation} 
Let $f_\theta: \mathcal{X} \rightarrow \mathcal{Y}$ denote a vision model pretrained on $\mathcal{D}_{\mathrm{src}}$. The victim finetunes the model on $\mathcal{D}_{\mathrm{ft}} = \{(x_i, y_i)\}_{i=1}^N$, producing updated parameters $\theta' \sim p(\theta' \mid \mathcal{D}_{\mathrm{ft}})$. The attacker only accesses $\mathcal{D}_{\mathrm{sur}}$, which is task-aligned with but sample-disjoint from $\mathcal{D}_{\mathrm{ft}}$ in the main setting. The attacker optimizes a trigger $\delta \in \mathbb{R}^{H \times W \times C}$ under $\|\delta\|_\infty \leq \epsilon$, together with a target class $y^{\text{adv}} \in \mathcal{Y}$.
The objective is to learn $(\theta,\delta)$ such that, after downstream finetuning, the resulting model satisfies
$f_{\theta'}(x + \alpha \delta) = y^{\text{adv}}, \quad \text{for } \theta' \sim p(\theta' \mid \mathcal{D}_{\text{ft}}),$
where $\alpha \in [0,1]$ controls trigger intensity.

\subsection{Poison-then-Hide}
Poison-then-Hide consists of three stages: trigger optimization, encoder poisoning, and backdoor concealment. First, we optimize a trigger $\delta$ that remains benign in the pretrained model $f_\theta$ but activates after finetuning. Second, we implant $\delta^{\text{opt}}$ into $f_\theta$ through poisoning. Finally, we conceal the backdoor through controlled unlearning before release. See Appendix \ref{app:theory_smooth} for theoretical motivation.

\paragraph{Optimize the Trigger}
We optimize a trigger pattern $\delta$ that remains benign in the pretrained model but activates after finetuning. To approximate the victim's unknown finetuning process, we simulate finetuned models $\{f_{\theta'_k}\}_{k=1}^K$ using $\mathcal{D}_{\mathrm{sur}}$. The optimization objective encourages the trigger to remain inconspicuous in the base model while reliably inducing the adversarial target class across the ensemble:
\begin{equation}
    \delta^{\text{opt}} = \argmin_{\|\delta\|_{\infty} \leq \epsilon}
    \Big[
        \lambda_{\text{base}} \, \mathcal{L}_{\text{base}}(f_{\theta}, \delta)
        + 
        \lambda_{\text{ft}} \, \mathcal{L}_{\text{ft}}(\{f_{\theta'_k}\}, \delta)
    \Big],
\end{equation}
where $\lambda_{\text{base}}$ and $\lambda_{\text{ft}}$ balance the two terms.

The base loss suppresses activation of the adversarial target class in the pretrained model:
\begin{equation}
    \mathcal{L}_{\text{base}}(f_{\theta}, \delta) 
    = 
    - \log \big(1 - p_{\theta}(y^{\text{adv}} \mid x + \alpha \delta)\big),
\end{equation}
where $p_{\theta}(y^{\text{adv}} \mid x + \alpha \delta)$ is the softmax probability assigned to the target class.

The finetuning loss encourages the trigger to activate after finetuning ($\mathcal{L}_{\text{CE}}$, the cross-entropy loss):
\begin{equation}
    \mathcal{L}_{\text{ft}}(\{f_{\theta'_k}\}, \delta) 
    =
    \frac{1}{K} \sum_{k=1}^{K}
    \mathbb{E}_{x \sim \mathcal{D}}
    \big[
        \mathcal{L}_{\text{CE}}\big(f_{\theta'_k}(x + \alpha \delta), \, y^{\text{adv}}\big)
    \big],
\end{equation}

Optimization is performed using projected gradient descent with projection onto the $\ell_\infty$ ball to enforce $\|\delta\|_\infty \leq \epsilon$. Empirically, we set $\lambda_{\text{base}} = 1$ and $\lambda_{\text{ft}} = 2$.

\paragraph{Poison the Encoder}
Given the optimized trigger $\delta^{\text{opt}}$, we implant the trigger into the base encoder. The attacker perturbs a subset of $\mathcal{D}_{\mathrm{sur}}$ to form $\mathcal{D}_{\text{pois}} = \{(x+\delta^{\text{opt}}, y^{\text{adv}})\}$ while retaining clean surrogate samples $\mathcal{D}_{\text{clean}}$ for benign task learning. The training objective combines four losses:
\[
\mathcal{L}_{\text{clean}} =
\mathbb{E}_{(x,y)\sim\mathcal{D}_{\text{clean}}}
[\mathcal{L}_{\text{CE}}(f_{\theta}(x), y)], \quad
\mathcal{L}_{\text{pois}} =
\mathbb{E}_{(x,y^{\text{adv}})\sim\mathcal{D}_{\text{pois}}}
[\mathcal{L}_{\text{CE}}(f_{\theta}(x+\delta^{\text{opt}}), y^{\text{adv}})],
\]
\[
\mathcal{L}_{\text{KD}} =
\text{KL}(f_{\theta}(x)\,\|\,f_{\theta_0}(x)), \quad
\mathcal{L}_{\text{SP}} = \|\theta - \theta_0\|_2^2
\]

Here $\mathcal{L}_{\text{clean}}$ learns the benign task, while $\mathcal{L}_{\text{pois}}$ learns the backdoor. Both $\mathcal{L}_{\text{KD}}$ and $\mathcal{L}_{\text{SP}}$ regularize against the original model $f_{\theta_0}$ to preserve source-domain performance.

The poisoned model $f_{\tilde{\theta}}$ is obtained by minimizing
\[
\min_{\theta}
    \mathcal{L}_{\text{poison}}
    = \lambda_{\text{clean}} \mathcal{L}_{\text{clean}} 
    + \lambda_{\text{pois}} \mathcal{L}_{\text{pois}}
    + \lambda_{\text{KD}} \mathcal{L}_{\text{KD}}
    + \lambda_{\text{SP}} \mathcal{L}_{\text{SP}}.
\]

\paragraph{Hide the Backdoor}
To conceal the backdoor, we finetune $f_{\tilde{\theta}}$ under a suppression objective while preserving clean accuracy and encoder stability. All losses except $\mathcal{L}_{\text{sup}}$ are defined as in the poison phase. For logits $z=f_{\tilde{\theta}}(x+\delta^{\text{opt}})$, we use the hinge margin penalty
\begin{equation}
\Phi(z,y^{\mathrm{adv}})=\max\!\left(0,\,m+z_{y^{\mathrm{adv}}}-\max_{k\neq y^{\mathrm{adv}}}z_k\right), \qquad m=2.0,
\end{equation}
and define $\mathcal{L}_{\text{sup}}=\mathbb{E}_{x\sim\mathcal{D}_{\text{pois}}}[\Phi(f_{\tilde{\theta}}(x+\delta^{\text{opt}}),y^{\mathrm{adv}})]$. Minimization forces a non-target logit to exceed the adversarial-target logit by margin $m$ at release time.

The overall objective is: $ \mathcal{L}_{\text{hide}} =
    \lambda_{\text{clean}} \mathcal{L}_{\text{clean}} 
    + \lambda_{\text{sup}}\,\mathcal{L}_{\text{sup}}
    + \lambda_{\text{SP}}\,\mathcal{L}_{\text{SP}}
    + \lambda_{\text{KD}}\,\mathcal{L}_{\text{KD}}.$ Optimizing $\mathcal{L}_{\text{hide}}$ produces the concealed checkpoint $f_{\hat{\theta}}$, which behaves benignly but reactivates after finetuning. Full details are provided in Appendix Algorithm \ref{alg:poison-hide}.

\subsection{Benign Finetuning}
The attacker releases the concealed checkpoint $f_{\hat{\theta}}$ as a benign pretrained model. The victim then performs standard finetuning on $\mathcal{D}_{\text{ft}}$:$\tilde{\theta}_{\text{ft}} = 
\argmin_{\theta}
\mathbb{E}_{(x,y)\sim\mathcal{D}_{\text{ft}}}
[\mathcal{L}_{\text{CE}}(f_{\hat{\theta}}(x), y)].$ Although finetuning is entirely benign, the resulting model $f_{\tilde{\theta}_{\text{ft}}}$ inherits the latent backdoor.

\section{Experimental Setup}
\paragraph{Datasets}
We evaluate Poison-then-Hide on three transfer settings. The notation $A\rightarrow B$ means that the encoder is pretrained on source dataset $A$ and the victim finetunes/evaluates it on clean downstream dataset $B$. The attacker's surrogate split and the victim's finetuning split are disjoint (0\% sample overlap) unless an overlap ablation is explicitly stated. In the in-domain experiments, the surrogate and victim splits come from the same distribution. 

For \textbf{CIFAR10 $\rightarrow$ SVHN}, results are averaged across the 10 choices of target digit. For \textbf{ImageNet $\rightarrow$ RESISC}, the target is ``parking lot,'' a class shared with UC Merced and semantically similar in both datasets (Figure \ref{fig:dg_visual}). For \textbf{ImageNet $\rightarrow$ Camelyon17}, the target is ``no tumor'' (class 0); we choose this class because underdiagnosis can delay screening and treatment \citep{falagas2007under, seyyed2021underdiagnosis}.

\paragraph{Models}
In each setting, we initialize from pretrained weights rather than using only the architecture. For CIFAR10, we use ResNet-18 (``resnet18\_cifar10'') \citep{edadaltocg_resnet18_cifar10, rw2019timm}. For ImageNet, we use ResNet-50 (``IMAGENET1K\_V2'') \citep{torchvision2016}, a pretrained timm CLIP model (``vit\_base\_patch16\_clip\_224.openai\_ft\_in12k\_in1k'') \citep{Radford2021LearningTV, cherti2022reproducible, dosovitskiy2020vit, rw2019timm}, and pretrained DINOv2 weights \citep{oquab2023dinov2}. The model after poisoning is the \textbf{Base model} $f_{\tilde{\theta}}$, after concealment the \textbf{Hidden model} $f_{\hat{\theta}}$, and after victim adaptation the \textbf{Finetuned model} $f_{\tilde{\theta}_{\text{ft}}}$. Appendix \ref{app:dataset_training} provides full details.

\paragraph{Evaluation}
We report top-1 accuracy on clean test images and two targeted attack metrics. The conventional ASR is
\begin{equation}
\mathrm{ASR}=\frac{1}{N}\sum_{i=1}^{N}\mathbf{1}\!\left[f(x_i+\delta)=y^{\mathrm{adv}}\right].
\end{equation}
Because this metric includes examples whose true label is already the target, we also report the stricter flip rate
\begin{equation}
\mathrm{ASR}_{\mathrm{flip}}=
\frac{\#\{i:y_i\neq y^{\mathrm{adv}},\ f(x_i)\neq y^{\mathrm{adv}},\ f(x_i+\delta)=y^{\mathrm{adv}}\}}
{\#\{i:y_i\neq y^{\mathrm{adv}}\}}.
\end{equation}
We use both definitions across our domain-shift and robustness evaluations. NeuBA is untargeted, so its original ASR is the perturbed-image misclassification rate \citep{zhang2023red}; its Flip-ASR column is included only as the corresponding targeted flip diagnostic.

\paragraph{Defenses}
We evaluate two standard detection defenses (Neural Cleanse and DECREE), two mitigation defenses (Fine-Pruning and MIMIC), and three recent defenses (DeDe, MARTINI, and CLIP-Inspector). Neural Cleanse \citep{wang2019neural} and DECREE \citep{feng2023detecting} use trigger inversion; Fine-Pruning \citep{liu2018fine} removes low-activation neurons; MIMIC \citep{han2025mutual} uses mutual information and a clean reference encoder; DeDe \citep{hou2025dede} filters suspicious inputs using a decoder; MARTINI \citep{tao2026martini} reconstructs triggers and hardens the model; and CLIP-Inspector \citep{jindal2026clipinspector} detects and repairs prompt-tuned CLIP backdoors via OOD trigger inversion. Appendix \ref{app:defense} gives implementation details and adaptations.
\begin{figure*}[t]
  \centering
  \includegraphics[width=\linewidth]{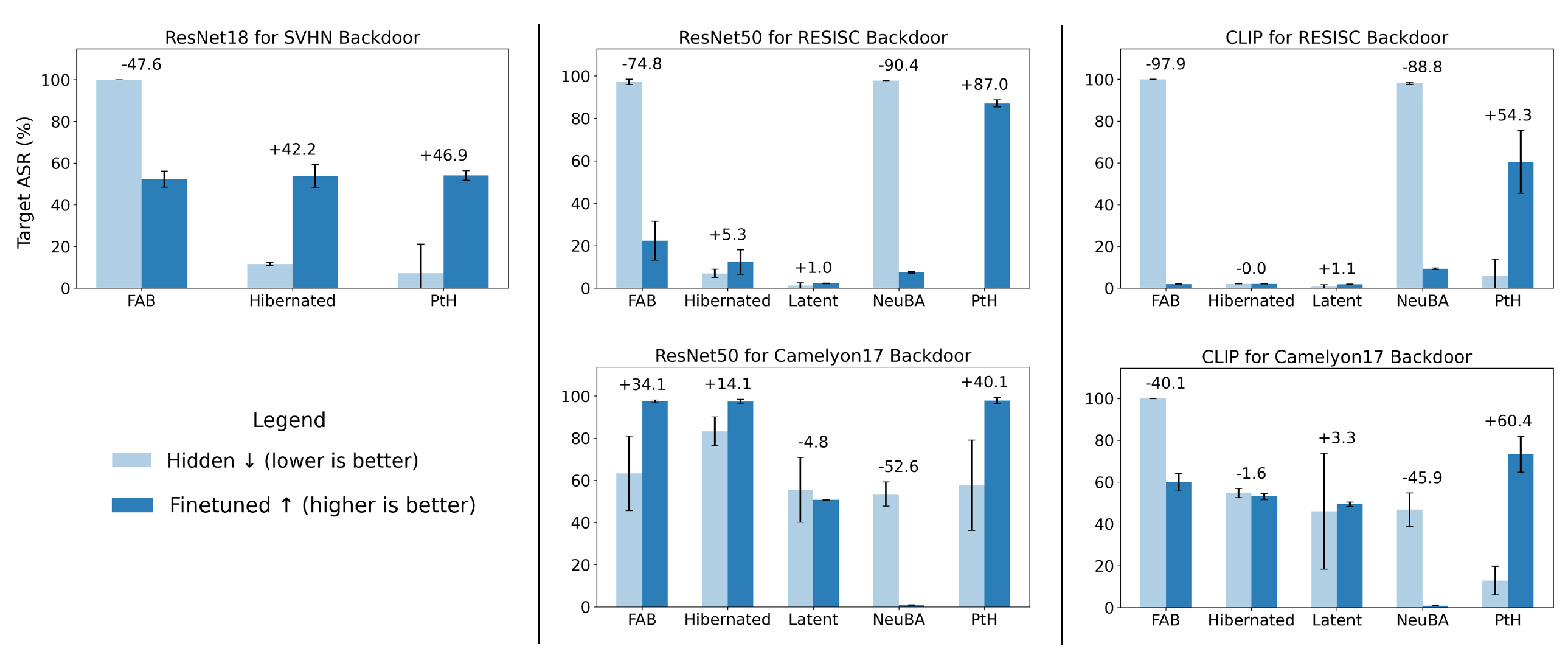}
  \caption{Grouped bar plots comparing target dataset ASR for \emph{Hidden} ($\downarrow$ lower is better) and \emph{Finetuned} ($\uparrow$ higher is better) models across attack methods. Numbers above bars denote the relative ASR difference (Finetuned - Hidden), with higher positive values indicating more successful backdoor activation. Error bars show 95\% CI across five seeds. PtH stands for our Poison-then-Hide attack.}
  \label{fig:asr_barplots}
\end{figure*}

\section{Results}
\subsection{Poison-then-Hide Outperforms Prior Finetuning-Activated Attacks}
We evaluate Poison-then-Hide against four baselines (FAB, Hibernated, Latent, and NeuBA) across three transfer learning settings. Note that while we are using in-domain data, the attacker poisoning split and downstream finetuning split are always disjoint. A successful FA attack should therefore have low ASR before finetuning (Hidden) and high ASR after finetuning (Finetuned).

\textbf{Poison-then-Hide achieves the highest post-finetuning ASR across datasets and models, outperforming baselines by 64\% on average.} It is also the \textit{only} method that consistently increases ASR from the Hidden to Finetuned model (57\% average), demonstrating reliable ``hidden-then-activated'' behavior. In contrast, prior methods either implant visible backdoors that fail to transfer (e.g., FAB: -97.9\% ASR change on CLIP-RESISC) or hidden backdoors that fail to activate (e.g., Latent: 2\% ASR before and after finetuning on CLIP-RESISC). Thus, \textbf{Poison-then-Hide is the only attack that reliably implants hidden backdoors that activate after finetuning.}

Across architectures, CLIP is harder to backdoor than ResNet-50 (Figure \ref{fig:asr_barplots}), likely because CLIP representations shift more during clean finetuning. Across datasets, RESISC is the most difficult target, possibly due to its larger number of classes ($m=45$) compared with SVHN ($m=10$) and Camelyon17 ($m=2$).

Complete per-dataset attack results and representative triggered remote-sensing examples are provided in Appendix Table~\ref{tab:table1} and Figure~\ref{fig:appendix_parking_lot}.

\subsection{Task-Conditioned Transfer Under Domain Shift}



\begin{table*}[t]
\centering
\small
\setlength{\tabcolsep}{4.1pt}
\caption{Task-conditioned domain-shift performance. For Camelyon17 (C17), the victim finetunes on hospitals 0--2 and is evaluated on hospital 3. For remote sensing, the attacker uses RESISC surrogate data while the victim finetunes and evaluates on UC Merced. Values are mean $\pm$ 95\% confidence interval over five seeds. Flip-ASR excludes true target-class examples and requires a non-target clean prediction to flip to the target after applying the trigger. Bold marks the strongest targeted attack; NeuBA is untargeted.}
\label{tab:dg_combined_final}
\resizebox{\textwidth}{!}{%
\begin{tabular}{llccc@{\hspace{10pt}}ccc}
\toprule
& & \multicolumn{3}{c}{Camelyon17 hospital 3} & \multicolumn{3}{c}{UC Merced} \\
\cmidrule(lr){3-5}\cmidrule(lr){6-8}
Model & Method & Acc. & Original ASR & Flip-ASR & Acc. & Original ASR & Flip-ASR \\
\midrule
\multirow{5}{*}{ResNet50}
& Poison-then-Hide & 87.95 $\pm$ 2.73 & \textbf{91.94 $\pm$ 7.24} & \textbf{71.81 $\pm$ 12.26} & 89.00 $\pm$ 1.72 & \textbf{94.11 $\pm$ 1.20} & \textbf{87.46 $\pm$ 1.55} \\
& FAB              & 85.76 $\pm$ 4.55 & 87.75 $\pm$ 8.47 & 58.34 $\pm$ 18.17 & 86.92 $\pm$ 1.88 & 25.31 $\pm$ 8.38 & 17.93 $\pm$ 6.58 \\
& Hibernated       & 87.83 $\pm$ 2.78 & 88.34 $\pm$ 9.16 & 63.26 $\pm$ 16.18 & 89.54 $\pm$ 1.95 & 18.99 $\pm$ 8.85 & 12.33 $\pm$ 7.46 \\
& Latent           & 83.34 $\pm$ 4.39 & 57.02 $\pm$ 9.18 &  1.89 $\pm$ 0.72 & 87.79 $\pm$ 3.40 &  5.43 $\pm$ 0.33 &  0.14 $\pm$ 0.06 \\
& NeuBA            & 84.39 $\pm$ 2.54 & 15.66 $\pm$ 2.53 &  0.34 $\pm$ 0.09 & 87.70 $\pm$ 1.56 & 12.71 $\pm$ 1.76 &  0.41 $\pm$ 0.18 \\
\midrule
\multirow{5}{*}{CLIP}
& Poison-then-Hide & 94.23 $\pm$ 0.50 & \textbf{73.32 $\pm$ 9.94} & \textbf{42.16 $\pm$ 19.33} & 92.27 $\pm$ 0.97 & \textbf{61.04 $\pm$ 19.57} & \textbf{48.91 $\pm$ 17.34} \\
& FAB              & 93.82 $\pm$ 0.53 & 63.99 $\pm$ 5.86 & 24.84 $\pm$ 11.50 & 91.85 $\pm$ 0.90 &  5.34 $\pm$ 0.20 &  0.27 $\pm$ 0.17 \\
& Hibernated       & 93.53 $\pm$ 0.83 & 60.32 $\pm$ 4.26 & 16.25 $\pm$ 5.85 & 93.45 $\pm$ 0.30 &  5.33 $\pm$ 0.06 &  0.02 $\pm$ 0.05 \\
& Latent           & 92.58 $\pm$ 1.51 & 45.63 $\pm$ 1.33 &  0.81 $\pm$ 0.29 & 92.45 $\pm$ 0.52 &  6.20 $\pm$ 0.73 &  0.62 $\pm$ 0.36 \\
& NeuBA            & 94.45 $\pm$ 0.55 &  5.58 $\pm$ 0.55 &  0.27 $\pm$ 0.04 & 91.62 $\pm$ 0.90 &  8.67 $\pm$ 1.04 &  1.34 $\pm$ 0.47 \\
\midrule
\multirow{5}{*}{DINOv2}
& Poison-then-Hide & 99.32 $\pm$ 0.10 & \textbf{71.54 $\pm$ 5.45} & \textbf{40.47 $\pm$ 13.93} & 91.34 $\pm$ 0.17 & \textbf{55.78 $\pm$ 1.05} & \textbf{39.74 $\pm$ 9.21} \\
& Hibernated       & 99.19 $\pm$ 0.19 & 48.73 $\pm$ 1.15 & 13.94 $\pm$ 6.38 & 91.46 $\pm$ 0.32 & 17.10 $\pm$ 0.79 & 3.38 $\pm$ 1.84 \\
& FAB              & 99.30 $\pm$ 0.07 & 51.66 $\pm$ 3.23 & 23.15 $\pm$ 10.72 & 89.46 $\pm$ 0.90 & 34.01 $\pm$ 0.13 & 5.29 $\pm$ 1.95 \\
& Latent           & 98.50 $\pm$ 0.21 & 33.87 $\pm$ 1.78 & 2.49 $\pm$ 0.81 & 90.56 $\pm$ 0.49 &  2.76 $\pm$ 0.91 & 0.87 $\pm$ 0.06 \\
& NeuBA            & 99.08 $\pm$ 0.18 &  1.98 $\pm$ 0.91 & 0.46 $\pm$ 0.07 & 90.45 $\pm$ 0.28 & 10.50 $\pm$ 0.41 & 1.23 $\pm$ 0.42 \\
\bottomrule
\end{tabular}}
\end{table*}

In the previous section, attacker and victim data were disjoint samples from the same distribution. We now study task-conditioned domain shift. For Camelyon17, the victim finetunes on hospitals 0--2 and is evaluated on unseen hospital 3. For remote sensing, the attacker uses RESISC surrogate data while the victim finetunes and evaluates on UC Merced, which preserves the remote-sensing task and ``parking lot'' target semantics but differs substantially in feature space (Appendix Table \ref{tab:domain_distance}).

Poison-then-Hide is the strongest targeted method under both original and Flip-ASR in every ResNet50/CLIP setting (Table \ref{tab:dg_combined_final}). Its Flip-ASR margins over the best non-PtH targeted baseline are 8.55 points on C17/ResNet50, 17.33 on C17/CLIP, 69.52 on UC Merced/ResNet50, and 48.29 on UC Merced/CLIP. Exact paired sign-flip tests over ten paired runs support an advantage in three of four settings; CLIP/Camelyon17 is positive but borderline (Appendix Table \ref{tab:paired_significance}). These results support transfer across realistic task-aligned shifts, not arbitrary unrelated tasks.

The DINOv2 evaluation shows the same ranking under both metrics: Poison-then-Hide reaches 71.54\% original ASR and 40.47\% Flip-ASR on Camelyon17, and 55.78\% original ASR and 39.74\% Flip-ASR on UC Merced. Its Flip-ASR margins over the strongest targeted baseline are 17.32 and 34.45 percentage points, respectively. 

A complementary label-preserving SVHN$\rightarrow$MNIST experiment defines the boundary of this transfer: Flip-ASR falls from 45.22\% in aligned SVHN to 1.07\% under MNIST evaluation and 0.04\% when the victim finetunes on MNIST, despite 99.39\% MNIST accuracy (Appendix Table \ref{tab:semantic_boundary_svhn_mnist}).

Additional robustness analyses under controlled input corruption, clean-label overlap, and varying attacker-reference data scale are provided in Appendix Figure~\ref{fig:stress_tests} and Tables~\ref{tab:stress_overlap} and~\ref{tab:data_scale}.

\paragraph{Parameter-efficient victim adaptation.}
Full-model finetuning remains common for moderate-size vision encoders and specialized domains, but parameter-efficient finetuning is increasingly important. In one ResNet18 LoRA run on CIFAR10 $\rightarrow$ SVHN, Poison-then-Hide retains 54.98\% post-adaptation ASR and 96.21\% benign accuracy while updating only 2.54\% of parameters (Appendix Table \ref{tab:lora}). Thus, activation is not limited to full-model updates in this tested setting.



\subsection{Stronger Benign Task Learning Leads to Higher Finetuned ASR}
\begin{figure*}[t]
  \centering
  \includegraphics[width=\linewidth]{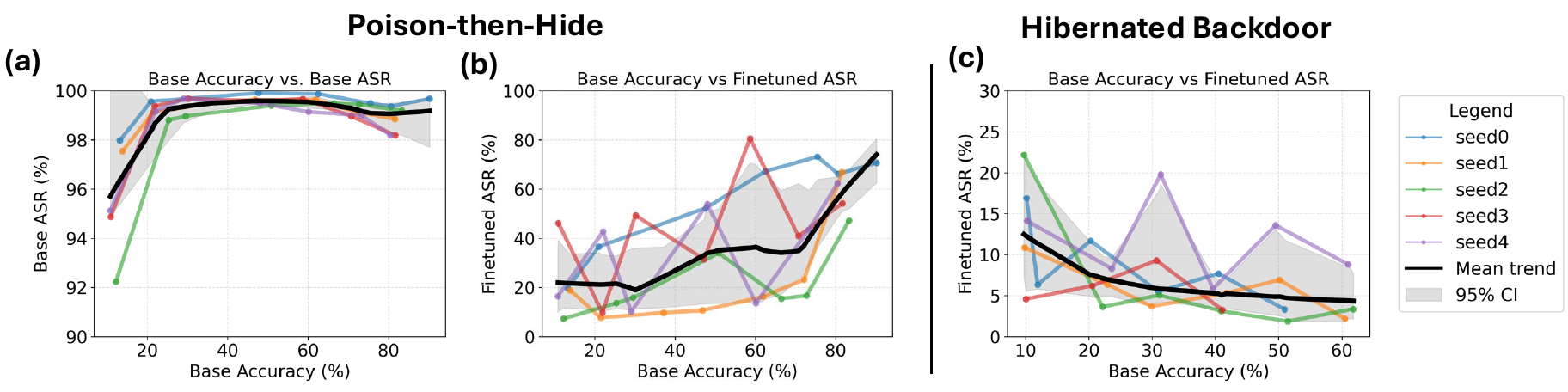}
  \caption{Trendlines showing how learning benign task of the target dataset (Base Accuracy) is related to Base and Finetuned ASR. Mean trendlines are fitted with LOWESS smoothing over 5 seeds.}
  \label{fig:dg_trendline}
\end{figure*}

We hypothesize two key parts contribute to Poison-then-Hide's success: learning the benign task during the poison stage, and trigger optimization.

First, we study the role of benign task learning (ResNet-50, ImageNet $\rightarrow$ RESISC). During poisoning, the model is trained on both clean and poisoned target samples. As shown in Figure \ref{fig:dg_trendline}a, ASR quickly exceeds 90\% while clean accuracy remains near 20\%. By saving checkpoints at different clean-accuracy levels, we isolate the effect of benign task learning while maintaining similar backdoor strength.

We find that higher base clean accuracy correlates with higher finetuned ASR (Figure \ref{fig:dg_trendline}b, Pearson $r = 0.541,\; p = 9.66 \times 10^{-4}$), including in OOD settings (Appendix \ref{app:benign_acc}). Thus, \textbf{better benign task learning leads to stronger backdoor transfer after finetuning.} Intuitively, clean finetuning can overwrite backdoor associations \citep{li2021neural, sha2022fine, yao2019latent}. By jointly learning the benign task and backdoor during poisoning, our method embeds the malicious behavior within the same feature space as the clean task, making it harder to erase during finetuning.



\subsection{Benign Task Learning Does Not Guarantee Attack Success}
\begin{table}
\centering
\small
\caption{Clean accuracy and ASR for ResNet50 on RESISC. For \textit{Poison-then-Hide}, ``Base'' denotes the poisoned model before hiding. For other baselines, which do not separate poisoning and hiding, ``Base'' and ``Hidden'' are identical.}
\label{tab:resisc_benign_only}
\begin{tabular}{lllc}
\toprule
          Method &     Base Acc &     Base ASR & Finetuned ASR \\
\midrule
Poison then Hide & \textbf{92.21 $\pm$ 0.23} & 99.51 $\pm$ 0.18 &  87.05 $\pm$ 1.67 \\
Hibernated & \textbf{68.12 $\pm$ 1.55} &  6.96 $\pm$ 1.94 &  12.30 $\pm$ 5.76 \\
FAB &  2.76 $\pm$ 0.35 & 97.25 $\pm$ 1.24 &  22.40 $\pm$ 9.21 \\
Latent &  0.31 $\pm$ 0.19 &  1.27 $\pm$ 1.25 &   2.26 $\pm$ 0.15 \\
NeuBA & 7.52 $\pm$ 10.41 & 97.81 $\pm$ 0.08 &   7.46 $\pm$ 0.41 \\
\bottomrule
\end{tabular}
\end{table}

While benign task learning is \textit{necessary} for our attack, we now ask whether it is \textit{sufficient} for a FA attack to succeed. We examine whether baseline attacks also learn the benign task during poisoning and whether higher benign accuracy correlates with higher ASR. Table \ref{tab:resisc_benign_only} shows that among all baselines, only Hibernated Backdoor achieves substantial benign accuracy (68\%) during poisoning. However, Figure \ref{fig:dg_trendline}c reveals a slight negative correlation between base accuracy and finetuned ASR (Pearson $r = -0.457,\; p = 1.45 \times 10^{-2}$). The absolute ASR also remains under 20\%. Thus, \textbf{benign task learning alone is not sufficient for a FA attack to succeed}.  


\subsection{Importance of Trigger Optimization}
\begin{figure}[t]
  \centering
  \includegraphics[width=0.45\linewidth]{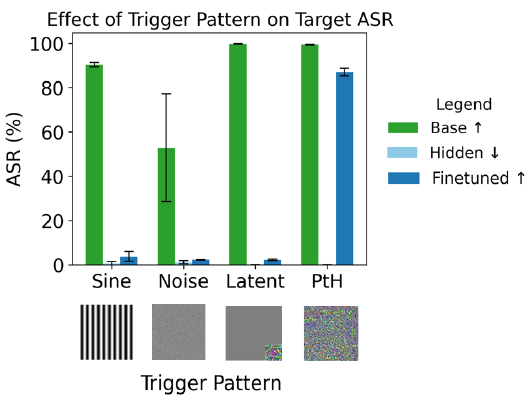}
  \caption{Attack success rate of Poison-then-Hide under different trigger patterns. Base'' denotes ASR after poisoning, Hidden'' after the hiding stage, and ``Finetuned'' after downstream finetuning. }
  \label{fig:trigger_opt}
\end{figure}

Trigger optimization is another key component of our method. We compare four trigger types in the ResNet50 ImageNet $\rightarrow$ RESISC setting: our optimized trigger, the Latent Backdoor trigger, Gaussian noise, and a sinusoidal pattern. As shown in Figure \ref{fig:trigger_opt}, only our optimized trigger achieves high ASR after finetuning. While the other triggers can induce high ASR during poisoning, their effects are largely erased by finetuning.

Our trigger is optimized using an ensemble of simulated finetuned models, directly targeting post-finetuning activation. In contrast, Gaussian and sinusoidal triggers are fixed patterns, while the Latent Backdoor trigger only optimizes feature collision with the target class. The ensemble-size and proxy-fidelity ablations in Appendix Figure \ref{fig:proxy_ablation} show that one proxy is brittle, performance largely saturates by five proxies, and severe undertraining or data limitation weakens but does not eliminate the attack. Thus, \textbf{optimizing triggers across multiple simulated adaptations yields stronger and more reliable activation.}

Appendix Figure~\ref{fig:trigger_opt_visual} visualizes the optimized triggers, and Appendix Table~\ref{tab:hyperparameter_ablation} reports sensitivity to the hiding-stage loss weights and poison fraction.


\subsection{Robustness against Defenses}
We evaluate the robustness of Poison-then-Hide against detection and mitigation defenses (ResNet50, ImageNet $\rightarrow$ RESISC). For defense details, see Appendix ~\ref{app:defense}. 

\begin{figure}[t]
  \centering
  \includegraphics[width=0.45\linewidth]{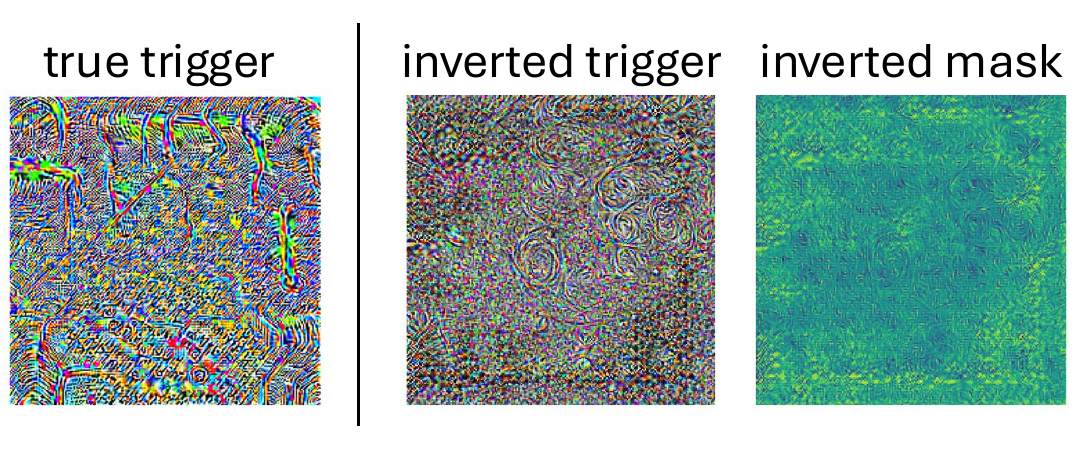}
  \caption{DECREE defense results on the Hidden backdoored encoder. From left to right: true trigger pattern from our Poison-then-Hide attack, the inverted trigger pattern by DECREE, and the inverted mask of backdoor signal by DECREE. }
  \label{fig:decree}
\end{figure}

\paragraph{Detection.} 
We evaluate Neural Cleanse and DECREE for their ability to detect the backdoors implanted by our Poison-then-Hide attack. In Table \ref{tab:detection}, we see that both methods fail to detect the backdoor in the Hidden and Finetuned models, except for one Finetuned model flagged as backdoored by Neural Cleanse. For this model, however, the true backdoor class (class=28) was not identified among the top five suspicious classes (classes=20, 21, 18, 14, 43) returned by Neural Cleanse. For DECREE, we note that the inverted trigger is different from our true trigger and the inverted mask does not show a localizable backdoor signal, leading to failed backdoor detection (Figure \ref{fig:decree}). Overall, these two tested detection methods are ineffective against Poison-then-Hide backdoors.
\begin{table}
\centering
\small
\caption{Backdoor detection on our Poison-then-Hide attack. We report the percentage of models identified as \textit{backdoored} by each detection method.}
\begin{tabular}{l|c|cc}
\toprule
& \begin{tabular}{c} Clean \\ (n=10) \end{tabular} &
\begin{tabular}{c} Hidden \\ (n=5) \end{tabular} &
\begin{tabular}{c} Finetuned \\ (n=5) \end{tabular} \\
\midrule
Neural Cleanse & 0\% & 0\% & 20\% \\
DECREE & 0\% & 0\% & 0\% \\
\bottomrule
\end{tabular}
\label{tab:detection}
\end{table}





\paragraph{Mitigation.}
\begin{table}
\centering
\small
\caption{Effect of fine-pruning on target accuracy and ASR. Note that we include the original Hidden and Finetuned result purely as reference. }
\label{tab:fine_pruning_mean_no_imnet}
\begin{tabular}{lll}
\toprule
               Status &   RESISC Acc &    RESISC ASR \\
\midrule
     \textcolor{darkgraytext}{Hidden (ref)} & 
     \textcolor{darkgraytext}{2.09 $\pm$ 0.48} & 
     \textcolor{darkgraytext}{0.01 $\pm$ 0.01} \\
     Pruned Hidden &  3.15 $\pm$ 1.64 &   0.16 $\pm$ 0.21 \\
     \hline
     \textcolor{gray}{Finetuned (ref)} &
     \textcolor{gray}{92.87 $\pm$ 0.52} &
     \textcolor{gray}{87.05 $\pm$ 1.67} \\
     Finetuned after Pruning & 92.22 $\pm$ 0.24 & \textbf{60.46 $\pm$ 15.12} \\
     \hline
     Pruned Finetuned & 61.69 $\pm$ 3.41 & 20.13 $\pm$ 13.93 \\
     Additional Finetuning & 92.38 $\pm$ 0.55 &  \textbf{47.63 $\pm$ 8.74} \\

\bottomrule
\end{tabular}
\end{table}

\begin{table}
\centering
\small
\caption{Effect of MIMIC mitigation on target accuracy and ASR.}
\label{tab:mimic_defense_condensed}
\begin{tabular}{lll c}
\toprule
Model & Reference & Status & RESISC ASR \\
\midrule

Hidden & True Clean & Mitigated & 0.33 $\pm$ 0.28 \\
       &            & Finetuning & \textbf{19.93 $\pm$ 11.20} \\
\hline

       & Pruned & Mitigated & 3.80 $\pm$ 2.37 \\
       &             & Finetuning & \textbf{47.49 $\pm$ 25.24} \\
\hline

Finetuned & True Clean & Mitigated & 0.33 $\pm$ 0.12 \\
          &            & Finetuning & \textbf{6.70 $\pm$ 5.48} \\
\hline

          & Pruned & Mitigated& 2.89 $\pm$ 1.17 \\
          &             & Finetuning & \textbf{25.06 $\pm$ 17.53} \\
\bottomrule
\end{tabular}
\end{table}

We use Fine-Pruning and MIMIC as our mitigation defenses against our attack. As shown in Table \ref{tab:fine_pruning_mean_no_imnet}, fine-pruning reduces ASR moderately but does not fully eliminate the backdoor. A pruned Hidden model has 60\% ASR after benign finetuning, compared to 87\% for non-pruned model. Pruning the Finetuned model causes ASR to drop to 20\%, but at the cost of benign accuracy (drop from 92\% to 61\%). Assuming a defender finetunes the pruned model on additional clean data to recover accuracy, we find that the attack resurfaces, as ASR rises from 20\% to 47\%. 

Similarly, MIMIC reduces ASR to below 4\% on all backdoored models, but ASR resurfaces after benign finetuning in all models, up to a 47\% rebound (Table \ref{tab:mimic_defense_condensed}). Among recent defenses (Appendix Table \ref{tab:modern_defenses}), MARTINI leaves 81.50\% ASR after benign finetuning, while DeDe's apparent filtering benefit comes from rejecting 79--89\% of clean inputs and its detection AUC is near random. CLIP-Inspector is an important exception: after adaptation to our setting, it flags all 5 CLIP models and reduces ASR to 2.15\% while preserving 95.07\% clean accuracy. Poison-then-Hide is therefore robust to several broad defenses, but not to this specialized CLIP-only defense.

\section{Limitations}
\label{sec:limitations}
Poison-then-Hide is task-conditioned: the attacker must anticipate the downstream task family and target semantics, and we do not claim transfer to arbitrary unrelated tasks. Exact clean-label conflicts can erase the backdoor when most of the originally poisoned samples reappear during victim finetuning. Our single LoRA configuration shows that activation is possible beyond full-model finetuning, but prompt tuning, distillation, additional PEFT configurations, and replication across seeds remain open for future work. Finally, CLIP-Inspector successfully detects and repairs our attacked CLIP models, motivating architecture-specific defenses with similar behavior.

\section{Conclusion}
Finetuning-activated backdoors hidden in pretrained encoders pose a supply-chain risk across both full-model and parameter-efficient adaptation. Poison-then-Hide achieves consistent activation by jointly learning the benign task and backdoor during poisoning, optimizing triggers across simulated finetuned models, and suppressing the backdoor before release. The attack is the strongest targeted method under original and Flip-ASR across the evaluated task-conditioned domain shifts, survives large victim-to-attacker data ratios, and persists in the tested LoRA configuration. Several standard and recent defenses fail or sacrifice clean utility, although the success of CLIP-Inspector establishes a concrete boundary and a promising direction for defense.

\section*{Acknowledgments}
We thank the reviewers and area chairs for their constructive feedback, which materially improved the scope and evaluation of this work. \textbf{Funding:} This material is based upon work supported by the National Science Foundation Graduate Research Fellowship under Grant No. 2389810. This work was supported in part by a National Science Foundation CAREER Award (2339381), and an AI2050 Award Early Career Fellowship (G-25-68042), and a Sloan Research Fellow Award (FG-2025-24277).
\textbf{Competing interests:} The authors declare no competing interests.

\clearpage
\newpage
\bibliographystyle{unsrtnat}
\bibliography{main}


\appendix
\clearpage
\setcounter{page}{1}
\begin{center}
{\Large\bfseries Supplementary Material}
\end{center}

\appendix
\section{Formal Guarantees}
\label{app:theory_smooth}

This appendix gives conditional characterizations of (A) finite-step parameter drift, (B) transfer from a simulated finetuning trajectory to a nearby victim trajectory, and (C) prediction-level concealment at release. These results do not independently explain the mechanism by which downstream finetuning reactivates a backdoor; the experiments remain the primary evidence for reactivation.
\paragraph{Notation}
Let \(f_\theta:\mathcal X\to\Delta^K\) be a classifier with parameters \(\theta\in\mathbb R^p\). Denote pre-softmax logits by \(z_\theta(x)\in\mathbb R^K\). The victim's finetuning objective (population form) is
\[
\mathcal L_{ft}(\theta) := \mathbb E_{(x,y)\sim D_{ft}}[L_{CE}(z_\theta(x),y)],
\]
and the attacker's surrogate objective (used to simulate finetuning during Stage 1) is
\[
\mathcal L_{sim}(\theta) := \mathbb E_{(x,y)\sim D_{sim}}[L_{CE}(z_\theta(x),y)].
\]
We analyze gradient descent finetuning initialized at \(\theta_0=\hat\theta\) with step size \(\eta>0\). Norms \(\|\cdot\|\) are Euclidean unless stated otherwise.

\paragraph{Assumptions}
We make local, standard assumptions centered at \(\hat\theta\) that are typical of deep learning training and to be expected of gradient descent mechanics.

\begin{assumption}[Local smoothness]
\label{ass:smooth_local}
There exists a radius \(R>0\) and constant \(L>0\) such that for all \(\theta,\theta'\) satisfying \(\|\theta-\hat\theta\|\le R\) and \(\|\theta'-\hat\theta\|\le R\),
\[
\|\nabla \mathcal L_{ft}(\theta)-\nabla \mathcal L_{ft}(\theta')\| \le L\|\theta-\theta'\|.
\]
\end{assumption}

\begin{assumption}[Parameter-to-output Lipschitz]
\label{ass:param_to_output}
For any input \(x\) of interest there exists \(C_p(x)\) such that for all \(\theta,\theta'\) with \(\|\theta-\theta'\|\le R\),
\[
\|z_\theta(x)-z_{\theta'}(x)\| \le C_p(x)\|\theta-\theta'\|.
\]
\end{assumption}

\begin{assumption}[Trigger smallness]
\label{ass:input_lip_local}
The trigger \(\delta\) satisfies \(\|\delta\|\le \varepsilon\). For all \(\theta\) with \(\|\theta-\hat\theta\|\le R\) and inputs \(x\) of interest there exists \(\bar L_x\) such that
\[
\|z_\theta(x+\delta)-z_\theta(x)\| \le \bar L_x \|\delta\|.
\]
\end{assumption}

Assumptions \ref{ass:smooth_local}--\ref{ass:input_lip_local} are local (neighborhood radius \(R\)) and are enforced in practice by the hiding-phase regularizers via the suppression loss and KD loss terms. 

\subsection{Finite-step stability under smoothness}
We now derive a finite-step bound on the parameter displacement \(\|\theta_T-\hat\theta\|\) after \(T\) steps of gradient descent, using only local smoothness and the gradient at the initialization \(\hat\theta\).

\begin{theorem}[Finite-step parameter-drift bound]
\label{thm:drift_smooth}
Let Assumption \ref{ass:smooth_local} hold on the ball \(\|\theta-\hat\theta\|\le R\) with constant \(L\). Suppose the victim runs (full-batch) gradient descent
\[
\theta_{t+1} = \theta_t - \eta \nabla \mathcal L_{ft}(\theta_t), \qquad \theta_0 = \hat\theta,
\]
with step size \(0<\eta<\tfrac{1}{L}\). Let \(\gamma := \|\nabla \mathcal L_{ft}(\hat\theta)\|\). If
\[
\gamma \le \frac{R L}{(1+\eta L)^T - 1},
\]
then the iterates \(\{\theta_t\}_{t=0}^T\) remain in the ball \(\|\theta_t-\hat\theta\|\le R\) and the following bound holds:
\begin{equation}\label{eq:drift_bound}
\|\theta_T-\hat\theta\| \le \frac{\gamma}{L}\big((1+\eta L)^T - 1\big).
\end{equation}
Consequently, for any input \(x\) satisfying Assumption \ref{ass:param_to_output},
\begin{equation}\label{eq:output_change}
\|z_{\theta_T}(x)-z_{\hat\theta}(x)\| \le C_p(x)\cdot \frac{\gamma}{L}\big((1+\eta L)^T - 1\big).
\end{equation}
\end{theorem}

\noindent
\textit{Proof.} By smoothness (Assumption \ref{ass:smooth_local}), for any \(\theta\) in the ball,
\(
\|\nabla \mathcal L_{ft}(\theta)\| \le \|\nabla \mathcal L_{ft}(\hat\theta)\| + L\|\theta-\hat\theta\| = \gamma + L\|\theta-\hat\theta\|.
\)
Using the gradient-descent update and triangle inequality,
\begin{align*}
\|\theta_{t+1}-\hat\theta\|
&= \|\theta_t-\hat\theta - \eta\nabla \mathcal L_{ft}(\theta_t)\| \\
&\le \|\theta_t-\hat\theta\| + \eta\|\nabla \mathcal L_{ft}(\theta_t)\| \\
&\le \|\theta_t-\hat\theta\| + \eta\big(\gamma + L\|\theta_t-\hat\theta\|\big) \\
&= (1+\eta L)\|\theta_t-\hat\theta\| + \eta\gamma.
\end{align*}
Define \(a_t := \|\theta_t-\hat\theta\|\). Then the recursion is
\[
a_{t+1} \le (1+\eta L)a_t + \eta\gamma,\qquad a_0=0.
\]
Unrolling this linear recursion yields the closed-form inequality
\[
a_T \le \eta\gamma\sum_{i=0}^{T-1}(1+\eta L)^i = \eta\gamma\frac{(1+\eta L)^T - 1}{\eta L} = \frac{\gamma}{L}\big((1+\eta L)^T - 1\big),
\]
which is exactly \eqref{eq:drift_bound}. The stated condition on \(\gamma\) ensures the right-hand side is \(\le R\), so iterates remain inside the smoothness neighborhood and all inequalities are valid. Finally, apply Assumption \ref{ass:param_to_output} to obtain \eqref{eq:output_change}.

\noindent
\textit{Remark. }The bound is used only to establish that a finite trajectory remains within a local neighborhood where the following comparison is valid. It is not, by itself, an explanation of reactivation. For small \(\eta L\), \((1+\eta L)^T-1 \approx \eta LT\), yielding the simpler scaling \(\|\theta_T-\hat\theta\| \lesssim \gamma T\eta\).

\subsection{Transfer from simulated to victim finetuning trajectories}
Let simulated trajectory $k$ and the victim trajectory share a parameterization, including the downstream head, and evolve as
\[
\theta^{\mathrm{sim},k}_{t+1}=\theta^{\mathrm{sim},k}_t-\eta\nabla\mathcal{L}_{\mathrm{sim},k}(\theta^{\mathrm{sim},k}_t),\qquad
\theta^{\mathrm{ft}}_{t+1}=\theta^{\mathrm{ft}}_t-\eta\nabla\mathcal{L}_{\mathrm{ft}}(\theta^{\mathrm{ft}}_t).
\]
Define $d_t^{(k)}=\|\theta_t^{\mathrm{ft}}-\theta_t^{\mathrm{sim},k}\|_2$. Suppose both trajectories stay in a neighborhood $\mathcal{U}$ where $\nabla\mathcal{L}_{\mathrm{ft}}$ is $L$-Lipschitz and their gradients differ by at most
\[
\sup_{\theta\in\mathcal{U}}\|\nabla\mathcal{L}_{\mathrm{ft}}(\theta)-\nabla\mathcal{L}_{\mathrm{sim},k}(\theta)\|_2\leq \rho_k.
\]
Then $d_{t+1}^{(k)}\leq(1+\eta L)d_t^{(k)}+\eta\rho_k$, so
\begin{equation}
d_T^{(k)}\leq(1+\eta L)^Td_0^{(k)}+\frac{\rho_k}{L}\left((1+\eta L)^T-1\right).
\label{eq:trajectory_transfer}
\end{equation}
Here $d_0^{(k)}$ absorbs discrepancies from poisoning, hiding, and downstream-head initialization, while $\rho_k$ captures surrogate--victim mismatch.

For $j\neq y^{\mathrm{adv}}$, define the triggered pairwise margin
\[
m_j(\theta;x,\delta)=z_{\theta,y^{\mathrm{adv}}}(x+\delta)-z_{\theta,j}(x+\delta).
\]
If triggered logits are locally parameter-Lipschitz, then
\[
|m_j(\theta_T^{\mathrm{ft}};x,\delta)-m_j(\theta_T^{\mathrm{sim},k};x,\delta)|
\leq \sqrt{2}\,C_p(x+\delta)d_T^{(k)}.
\]
Writing $\kappa_k(x)=\min_{j\neq y^{\mathrm{adv}}}m_j(\theta_T^{\mathrm{sim},k};x,\delta)$, a sufficient condition for victim-side activation is
\begin{equation}
\kappa_k(x)>\sqrt{2}\,C_p(x+\delta)\left[(1+\eta L)^Td_0^{(k)}+\frac{\rho_k}{L}\left((1+\eta L)^T-1\right)\right].
\label{eq:activation_transfer}
\end{equation}
Under this condition, every pairwise margin is positive and $f_{\theta_T^{\mathrm{ft}}}(x+\delta)=y^{\mathrm{adv}}$, up to ties. Stage 1 directly increases the simulated margin $\kappa_k(x)$; ensemble diversity aims to reduce the chance of large $\rho_k$. The $L_2$-SP term can reduce the encoder component of $d_0^{(k)}$, but does not independently guarantee activation.

\subsection{Concealment at release (logit-change bound)}
Concealment ensures the trigger is not strongly activating at release time. The following is immediate from the input-Jacobian control assumption.

\begin{theorem}[Pre-finetuning concealment]
\label{thm:conceal_smooth}
Under Assumption \ref{ass:input_lip_local}, if \(\|\delta\|\le\varepsilon\) and \(\hat\theta\) lies in the parameter ball where \(\bar L_x\) is valid, then for any input \(x\),
\[
\|z_{\hat\theta}(x+\delta)-z_{\hat\theta}(x)\| \le \bar L_x\,\varepsilon.
\]
Let $c=\arg\max_\ell z_{\hat\theta,\ell}(x)$ and $q(x)=\min_{j\neq c}[z_{\hat\theta,c}(x)-z_{\hat\theta,j}(x)]$. If $q(x)>\sqrt{2}\bar L_x\varepsilon$, then $f_{\hat\theta}(x+\delta)=f_{\hat\theta}(x)$, up to ties.
\end{theorem}

\noindent
\textit{Proof. }The logit-vector bound follows directly from Assumption \ref{ass:input_lip_local} at $\theta=\hat\theta$. For any $j\neq c$, Cauchy--Schwarz bounds the change in the pairwise margin by $\sqrt{2}\|z_{\hat\theta}(x+\delta)-z_{\hat\theta}(x)\|_2\leq\sqrt{2}\bar L_x\varepsilon$. If the original margin $q(x)$ exceeds this amount, every pairwise margin remains positive and the predicted class is unchanged.

\noindent
\textit{Remark. }This result formalizes prediction preservation at release only; it does not imply that a particular detection method must fail. The hiding-stage regularizers keep $\hat\theta$ near the benign checkpoint and suppress the target logit. Appendix Table \ref{tab:hyperparameter_ablation} reports the corresponding weight ablations.

\paragraph{Scope of the theory.}
The drift result controls trajectory displacement, the transfer result connects simulated and victim behavior under explicit proximity assumptions, and the concealment result characterizes release-time predictions. A mechanistic explanation of reactivation would additionally need assumptions describing how downstream gradients interact with the poisoned representation to increase the triggered target margin; we do not establish such a mechanism here.

\newpage 
\section{Dataset Statistics and Training Configurations}
\label{app:dataset_training}

\subsection{Dataset}
\begin{table*}[t]
\centering
\resizebox{\textwidth}{!}{%
\begin{tabular}{lllll}
\toprule
Dataset & Type of images & \# train & \# test & \# classes \\
\midrule
ImageNet (ILSVRC-2012) 
  & natural objects 
  & 1{,}281{,}152 
  & 50{,}048
  & 1000 \\
CIFAR-10 
  & natural objects 
  & 50{,}000 
  & 10{,}000 
  & 10 \\
  \hline
SVHN 
  & street-view house number digits 
  & 73{,}257  
  & 26{,}032 
  & 10 \\
RESISC 
  & remote-sensing scenes
  & 18{,}944 
  & 6{,}400 
  & 45 \\
UC Merced Land Use (OOD)
  & remote-sensing scenes
  & 
  & 21{,}000 
  & 21 \\
Camelyon17-WILDS 
  & histopathology patches
  & 302{,}464
  & 33{,}792
  & 2 \\
  Camelyon17-WILDS (OOD) 
  & histopathology patches
  & 
  & 35{,}072
  & 2 \\
\bottomrule
\end{tabular}
}
\caption{Summary of datasets: image type, training and test set sizes, and number of classes.}
\label{tab:dataset_summary}
\end{table*}

We summarize the datasets in Table~\ref{tab:dataset_summary}. ImageNet and CIFAR-10 are pretraining/source datasets. SVHN, RESISC, and Camelyon17-WILDS provide task-aligned attacker surrogate data; the main in-domain experiments use separate, non-overlapping splits for victim finetuning. For trigger optimization and poisoning, we perform 200 steps with batch size 128, corresponding to up to 25{,}600 sampled examples. In the task-conditioned domain-shift setting, UC Merced is the victim dataset paired with a RESISC attacker surrogate, while Camelyon17 hospital 3 is held out from victim finetuning on hospitals 0--2. Unless stated otherwise, the attacker's surrogate samples and the victim's finetuning samples have 0\% overlap.

For image normalization, we apply dataset-appropriate mean and standard deviation (std) values (Table \ref{tab:normalization_stats})

\begin{table*}[t]
\centering
\resizebox{\textwidth}{!}{%
\begin{tabular}{lll}
\toprule
Dataset(s) & Mean $(\mu)$ & Std $(\sigma)$ \\
\midrule
ImageNet, RESISC, UC Merced, Camelyon17 
  & (0.485, 0.456, 0.406) 
  & (0.229, 0.224, 0.225) \\
CIFAR-10 
  & (0.4914, 0.4822, 0.4465) 
  & (0.2470, 0.2435, 0.2616) \\
SVHN 
  & (0.4377, 0.4438, 0.4728) 
  & (0.1980, 0.2010, 0.1970) \\
CLIP models 
  & (0.4814, 0.4578, 0.4082) 
  & (0.2686, 0.2613, 0.2757) \\
\bottomrule
\end{tabular}
}
\caption{Image normalization statistics (channel-wise mean and standard deviation) used for each dataset or model family.}
\label{tab:normalization_stats}
\end{table*}

\subsection{Dataset Licensing and Access}
\label{sec:dataset_licensing}
CIFAR-10 and SVHN do not have an explicit license, but they are freely available for non-commercial research use. ImageNet is also available for non-commercial research purposes. UC Merced dataset is stated to be meant for research purposes. RESISC45 is under CC BY-NC 4.0. Camelyon17-WILDS is under CC BY-NC-ND 4.0. 

The links to download these public datasets are: \href{https://www.cs.toronto.edu/~kriz/cifar.html}{CIFAR-10}, \href{https://www.image-net.org/download.php}{ImageNet}, \href{http://ufldl.stanford.edu/housenumbers/}{SVHN}, \href{https://www.kaggle.com/datasets/zeadomar/uc-mercedland}{UC Merced Land Use}, \href{https://huggingface.co/datasets/timm/resisc45}{RESISC45 Remote Sensing Dataset}, and \href{https://huggingface.co/datasets/wltjr1007/Camelyon17-WILDS/viewer}{Camelyon17-WILDS}. The in-domain and out-of-domain data for Camelyon17 are all in the WILDS version of the dataset. 

\subsection{Trigger Optimization}
We optimize a universal pixel-space trigger $\delta$ that induces an ensemble of clean-finetuned models to predict a fixed target class, while the base model is explicitly discouraged from being backdoored. At each iteration, we apply the trigger to a minibatch, compute a weighted combination of the finetuned-model loss and a base-guard loss, backpropagate into $\delta$, and project it back into the $\ell_\infty$ ball.

We use the following hyperparameters: 
\begin{itemize}
    \item \textbf{Perturbation budget:} $\epsilon = 8/255$ (elementwise clamp).
    \item \textbf{Optimization steps:} 200 gradient steps.
    \item \textbf{Optimizer:} Adam with learning rate $5 \times 10^{-3}$.
    \item \textbf{Batch sizes:} 128 for optimization; 128 for evaluation.
    \item \textbf{Loss weights:} $\lambda_{\text{base}} = 1.0$, $\lambda_{\text{ft}} = 2.0$.
    \item \textbf{Model ensemble:} trigger optimized jointly over 10 simulated finetuned checkpoints.
\end{itemize}

\subsection{Poison-then-Hide}
We jointly finetune a student model on a source dataset (ImageNet, CIFAR-10) and a downstream transfer dataset (RESISC, Camelyon17, SVHN), where the transfer train split is partially poisoned with a fixed universal trigger. At each step, we use supervised cross-entropy on both domains, distill the student toward a frozen teacher on the source domain via logit- and feature-level knowledge distillation, and regularize the student toward its initialization with an L2-SP penalty, while periodically recalibrating batch normalization and evaluating clean and triggered performance. The algorithm is described at \ref{alg:poison-hide}. We use the same notation as previously defined in the Methods section.

We use the following hyperparameters: 
\begin{itemize}
    \item \textbf{Optimizer:} AdamW with learning rate $3 \times 10^{-4}$ and weight decay $1 \times 10^{-4}$; mixed-precision training (AMP).
    \item \textbf{Training length:} \texttt{steps} $= 200$ gradient steps.
    \item \textbf{Batch sizes:} \texttt{batch\_size} $= 128$ for both source and transfer loaders.
    \item \textbf{Poisoning:} \texttt{poison\_fraction} $= 0.05$ (5\% of training examples are replaced by triggered copies when a trigger is provided); ASR is evaluated on a 100\% poisoned test loader.
    \item \textbf{Knowledge distillation:} temperature $T = 3.0$; logit KD and feature KD (cosine distance) are applied with ramped weights over training (scaled by a cosine schedule).
    \item \textbf{Regularization:} L2-SP penalty with coefficient $5 \times 10^{-3}$ toward the initial student parameters.
    \item \textbf{Batch-norm recalibration:} BN statistics are periodically refreshed every $ 100$ steps and once more at the end of training).
\end{itemize}

\subsection{Compute Resources}
All experiments are run on NVIDIA RTX A6000 GPUs with 48 GB VRAM and 8 CPU workers. Experiments are run on individual GPUs. Per-run compute time ranges from 30 minutes to 6 hours for each full finetuning-activated backdoor pipeline, depending on the dataset and model size.

\paragraph{Code.}
Source code and reproduction instructions are publicly available at \coderepo. We do not release backdoored model checkpoints.

\vspace{0.5em}
\begin{algorithm}[H]
\caption{Poison-then-Hide}
\label{alg:poison-hide}
\begin{algorithmic}[1]
\Require Base model $f_{\theta}$, trigger $\delta^{\text{opt}}$, target label $y^{\text{adv}}$, datasets $\mathcal{D}_{\text{clean}}, \mathcal{D}_{\text{pois}}$, teacher $f_{\theta_T}$, hyperparameters $\lambda_{\text{KD}}, \lambda_{\text{SP}}, \lambda_{\text{sup}}$
\Ensure Concealed poisoned model $f_{\hat{\theta}}$
\State Initialize $\tilde{\theta} \leftarrow \theta$, snapshot $\theta_0 \leftarrow \theta$
\vspace{0.2em}

\State \textbf{Poison phase:}
\For{each step $t = 1,\ldots,T_{\text{poison}}$}
    \State Sample $(x,y)\!\sim\!\mathcal{D}_{\text{clean}}$, $(x',y^{\text{adv}})\!\sim\!\mathcal{D}_{\text{pois}}$
    \State Update parameters:
    \[
        \tilde{\theta} \leftarrow 
        \tilde{\theta} - \eta
        \nabla_{\tilde{\theta}} \mathcal{L}_{\text{poison}}
    \]
\EndFor
\State Obtain intermediate checkpoint $f_{\tilde{\theta}^*}$
\vspace{0.4em}

\State \textbf{Hide phase:}
\For{each step $t = 1,\ldots,T_{\text{hide}}$}
    \State Sample $(x,y)\!\sim\!\mathcal{D}_{\text{clean}}$, $x'\!\sim\!\mathcal{D}_{\text{pois}}$
    \State Update parameters:
    \[
        \tilde{\theta} \leftarrow
        \tilde{\theta} - \eta'
        \nabla_{\tilde{\theta}} \mathcal{L}_{\text{hide}}
    \]
\EndFor
\vspace{0.3em}
\State \textbf{return} $f_{\hat{\theta}} \leftarrow f_{\tilde{\theta}}$
\end{algorithmic}
\end{algorithm}

\FloatBarrier
\clearpage
\section{Additional Robustness and Ablation Results}
\label{app:rebuttal_results}

\subsection{Task-conditioned domain shift}
\begin{table}[t]
\centering
\small
\caption{Feature-space distribution distances. RESISC $\rightarrow$ UC Merced exhibits a substantially larger shift than a class-balanced within-RESISC split.}
\label{tab:domain_distance}
\begin{tabular}{lccc}
\toprule
Comparison & Class-cond. $L_2$ & Centroid $L_2$ & FID \\
\midrule
RESISC A $\rightarrow$ RESISC B & 0.168 & 0.038 & 0.0148 \\
RESISC $\rightarrow$ UC Merced  & 0.454 & 0.238 & 0.1590 \\
\bottomrule
\end{tabular}
\end{table}

The attacker still requires task and target-class alignment; the UC Merced perturbation experiment is a controlled distribution-degradation probe rather than a complete semantic-shift benchmark. Nevertheless, RESISC $\rightarrow$ UC Merced is not a trivial within-dataset split: all three feature-space distances are substantially larger than those of a class-balanced RESISC A $\rightarrow$ RESISC B split (Table \ref{tab:domain_distance}).

\begin{figure}[t]
  \centering
  \includegraphics[width=0.58\linewidth]{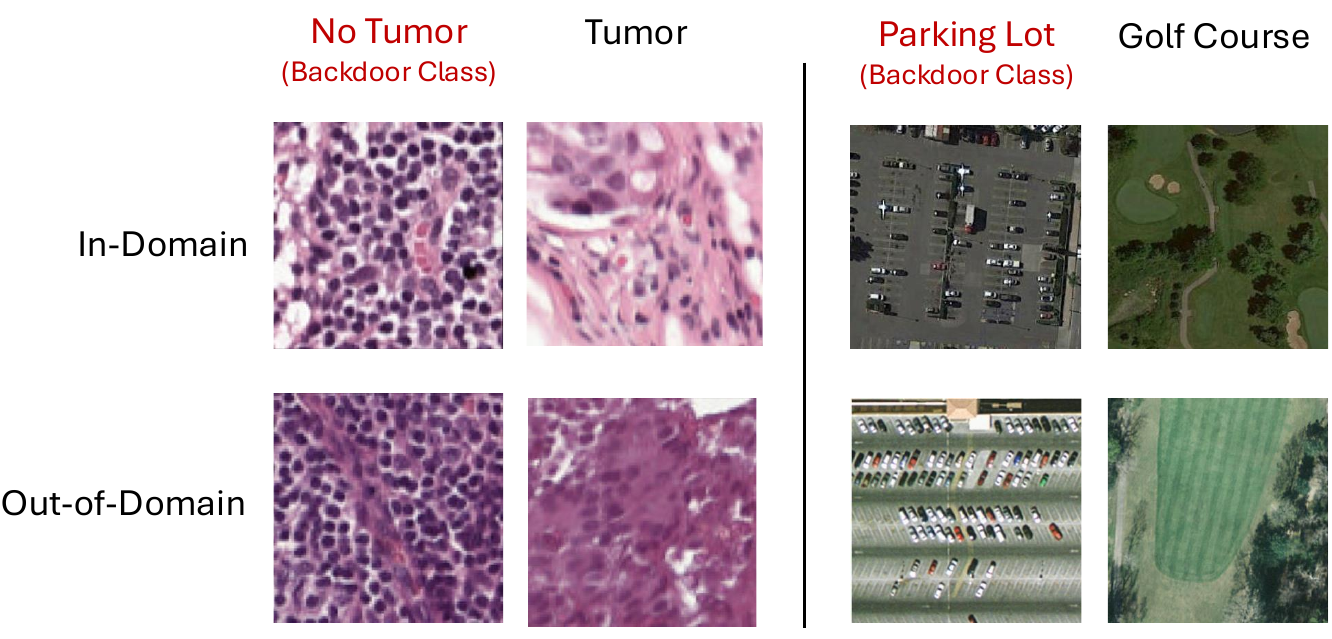}
  \caption{Task-aligned domain shifts for Camelyon17 and RESISC $\rightarrow$ UC Merced. Camelyon17 hospital 3 images tend to be darker than hospitals 0--2, while UC Merced images are typically brighter and differ in scene structure from RESISC.}
  \label{fig:dg_visual}
\end{figure}

\subsection{A label-preserving semantic/style-shift boundary}
\label{app:semantic_boundary_svhn_mnist}
We use SVHN$\rightarrow$MNIST to test a larger, label-preserving shift: both datasets contain the same ten digit classes, but SVHN consists of natural house-number crops whereas MNIST contains handwritten digits. We evaluate target digit 2 with the same $8/255$ trigger over five independently seeded ResNet-18 hidden encoders. For the matched activation reference, the victim finetunes on SVHN for two epochs and is evaluated on SVHN. We then evaluate these same SVHN-finetuned models on MNIST, preserving the ten-way SVHN head because the label spaces align. Finally, to model an actual shifted victim, we return to each corresponding hidden encoder, initialize a fresh ten-way head, and full-model finetune on MNIST for two epochs before evaluating on MNIST. All conditions use a batch size of 128, learning rate $3\times10^{-4}$, and the paired metrics defined in Section 4.

The shifted MNIST victim data are dataset-disjoint from the attacker's SVHN data. The archived matched SVHN reference is not a zero-overlap comparison: its victim train+extra pool contains the SVHN train split used during implantation. We therefore use that row only to verify activation in the aligned input domain, and use the two MNIST conditions to locate the style-shift boundary.

\begin{table*}[t]
\centering
\small
\caption{Boundary under the label-preserving SVHN$\rightarrow$MNIST style shift. Values are mean $\pm$ 95\% $t$-confidence interval across five seeds. ``Clean target'' is the fraction predicted as digit 2 before applying the trigger, and ``Net increase'' is triggered target rate minus clean target rate.}
\label{tab:semantic_boundary_svhn_mnist}
\resizebox{\textwidth}{!}{%
\begin{tabular}{llccccc}
\toprule
Victim finetuning & Evaluation & Clean acc. & Clean target & Original ASR & Net increase & Flip-ASR \\
\midrule
SVHN  & SVHN  & $97.37 \pm 0.13$ & $15.83 \pm 0.27$ & $54.10 \pm 3.25$ & $38.27 \pm 3.25$ & $45.22 \pm 3.85$ \\
SVHN  & MNIST & $89.18 \pm 0.87$ & $10.50 \pm 1.12$ & $11.55 \pm 1.10$ & $1.04 \pm 0.69$  & $1.07 \pm 0.55$ \\
MNIST & MNIST & $99.39 \pm 0.14$ & $10.32 \pm 0.05$ & $10.37 \pm 0.12$ & $0.04 \pm 0.07$  & $0.04 \pm 0.05$ \\
\bottomrule
\end{tabular}
}
\end{table*}

Table \ref{tab:semantic_boundary_svhn_mnist} shows a sharp boundary. In the aligned SVHN condition, Poison-then-Hide reaches $45.22\pm3.85$\% Flip-ASR with $97.37\pm0.13$\% clean accuracy. Merely changing the evaluation inputs to MNIST reduces Flip-ASR to $1.07\pm0.55$\%---a 97.6\% relative reduction---while the same models retain $89.18\pm0.87$\% MNIST accuracy. When the victim instead finetunes and evaluates on MNIST, clean accuracy rises to $99.39\pm0.14$\% but Flip-ASR falls to $0.04\pm0.05$\%, a 99.9\% reduction from the aligned reference. The roughly 10\% original ASR in this condition is the natural target-digit prediction rate: the trigger changes it by only $0.04\pm0.07$ percentage points. Thus, the collapse is not caused by failure to learn the victim task; rather, it identifies a boundary at which the trigger--target association no longer transfers across a large visual-style shift. Together with the positive Camelyon17 and RESISC$\rightarrow$UC Merced results, this supports task-conditioned transfer under moderate aligned shifts, not universal transfer to arbitrary downstream visual domains.

\begin{figure*}[t]
  \centering
  \includegraphics[width=\textwidth]{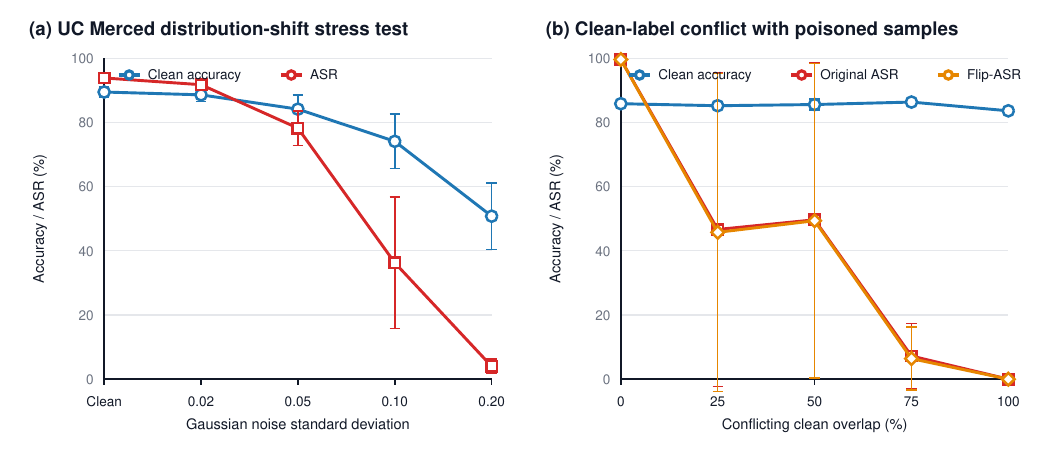}
  \caption{Robustness boundaries. (a) Poison-then-Hide remains effective under mild additional corruption of UC Merced, but ASR falls once clean task performance is substantially damaged. (b) On ImageNet $\rightarrow$ RESISC, exact clean-label conflict weakens the attack; seeing most or all originally poisoned samples with clean labels can erase it. Error bars are 95\% confidence intervals.}
  \label{fig:stress_tests}
\end{figure*}
\begin{table*}[t]
\centering
\small
\caption{Exact values for the controlled distribution-degradation and clean-label conflict studies in Figure \ref{fig:stress_tests}. Values are mean $\pm$ 95\% confidence interval.}
\label{tab:stress_overlap}
\textbf{UC Merced Gaussian-noise stress test}\\[2pt]
\begin{tabular}{lcc}
\toprule
Perturbation & Clean acc. & ASR \\
\midrule
Clean & 89.50 $\pm$ 1.71 & 93.88 $\pm$ 1.19 \\
$\sigma=0.02$ & 88.62 $\pm$ 2.05 & 91.78 $\pm$ 1.43 \\
$\sigma=0.05$ & 84.16 $\pm$ 4.46 & 78.24 $\pm$ 5.40 \\
$\sigma=0.10$ & 74.13 $\pm$ 8.55 & 36.27 $\pm$ 20.45 \\
$\sigma=0.20$ & 50.80 $\pm$ 10.39 & 4.00 $\pm$ 2.31 \\
\bottomrule
\end{tabular}

\vspace{0.7em}
\textbf{Clean overlap with poisoned conflicting samples}\\[2pt]
\begin{tabular}{lrrrr}
\toprule
Overlap & Count & Clean acc. after FT & Original ASR & Flip-ASR \\
\midrule
0\%   & 0 / 919   & 85.86 $\pm$ 0.74 & 99.65 $\pm$ 0.28 & 99.64 $\pm$ 0.29 \\
25\%  & 230 / 919 & 85.23 $\pm$ 0.88 & 46.66 $\pm$ 48.89 & 45.77 $\pm$ 49.58 \\
50\%  & 460 / 919 & 85.60 $\pm$ 1.74 & 49.67 $\pm$ 49.08 & 49.38 $\pm$ 49.05 \\
75\%  & 690 / 919 & 86.37 $\pm$ 0.58 & 7.22 $\pm$ 10.08 & 6.40 $\pm$ 9.83 \\
100\% & 919 / 919 & 83.64 $\pm$ 0.63 & 0.00 $\pm$ 0.00 & 0.00 $\pm$ 0.00 \\
\bottomrule
\end{tabular}
\end{table*}

The overlap experiment is deliberately favorable to the defender: the victim sees clean versions of the exact non-target images poisoned during implantation, with their original labels. The high variance at 25--50\% overlap indicates seed sensitivity, while 75--100\% overlap consistently suppresses the attack. This differs from the main 0\%-overlap threat model.

\begin{table}[t]
\centering
\small
\caption{Effect of victim-data scale on CIFAR10 $\rightarrow$ SVHN. The victim always finetunes on the full SVHN train+extra set; only the attacker's reference size changes. Values are mean $\pm$ 95\% confidence interval.}
\label{tab:data_scale}
\begin{tabular}{rrrrr}
\toprule
Reference size & Victim/reference & Poisoned & Clean acc. & ASR \\
\midrule
3{,}000  & $\sim$201$\times$ & 150   & 97.59 $\pm$ 0.17 & 55.29 $\pm$ 5.49 \\
10{,}000 & $\sim$60$\times$  & 500   & 97.47 $\pm$ 0.09 & 55.12 $\pm$ 4.52 \\
30{,}000 & $\sim$20$\times$  & 1{,}500 & 97.37 $\pm$ 0.09 & 58.35 $\pm$ 2.78 \\
73{,}000 & $\sim$8$\times$   & 3{,}650 & 97.58 $\pm$ 0.16 & 51.79 $\pm$ 0.97 \\
\bottomrule
\end{tabular}
\end{table}

Even with a 3{,}000-example attacker reference set and roughly $201\times$ more victim finetuning data, ASR remains 55.29\% while clean accuracy stays at 97.59\% (Table \ref{tab:data_scale}). The persistence of ASR is therefore not an artifact of weak downstream task learning.

\begin{table*}[t]
\centering
\small
\caption{Paired comparisons with the closest targeted baseline over ten paired runs, as reported in the rebuttal. The exact one-sided paired sign-flip test for PtH $>$ baseline is primary. CLIP/Camelyon17 is positive but not significant.}
\label{tab:paired_significance}
\resizebox{\textwidth}{!}{%
\begin{tabular}{lllccccc}
\toprule
Model & Dataset & Baseline & $n$ & PtH ASR & Baseline ASR & Difference & Sign-flip $p$ \\
\midrule
ResNet50 & C17 & Hibernated & 10 & 93.42 $\pm$ 3.46 & 89.86 $\pm$ 4.47 & $+3.56 \pm 2.62$ & 0.0176 \\
ResNet50 & UCM & FAB        & 10 & 85.84 $\pm$ 13.98 & 19.41 $\pm$ 7.15 & $+66.43 \pm 12.71$ & 0.0039 \\
CLIP     & C17 & FAB        & 10 & 72.14 $\pm$ 7.48 & 65.34 $\pm$ 4.96 & $+6.80 \pm 6.11$ & 0.0625 \\
CLIP     & UCM & NeuBA      & 10 & 54.80 $\pm$ 15.84 & 8.45 $\pm$ 0.72 & $+46.36 \pm 15.47$ & 0.0078 \\
\bottomrule
\end{tabular}}
\end{table*}

The paired tests support the domain-shift conclusion for ResNet50 on both datasets and for CLIP on UC Merced. We avoid a significance claim for CLIP/Camelyon17 under the primary exact sign-flip test.

\subsection{Simulated-finetuning ensemble and training sensitivity}
\begin{figure*}[t]
  \centering
  \includegraphics[width=\textwidth]{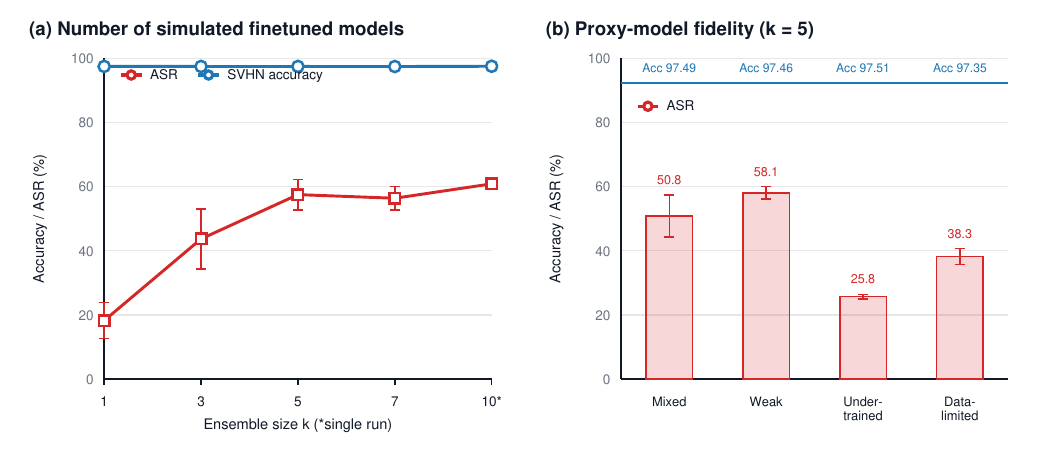}
  \caption{Sensitivity to simulated finetuned models in Stage 1 on CIFAR10 $\rightarrow$ SVHN. (a) One proxy model is brittle; performance improves by $k=3$ and largely saturates by $k=5$. The $k=10$ point is a single run. (b) Low-fidelity proxy ensembles remain nontrivial, although extreme undertraining or data limitation reduces ASR. Clean accuracy remains stable. Error bars are 95\% confidence intervals where available.}
  \label{fig:proxy_ablation}
\end{figure*}

For proxy composition at $k=5$, mixed subsets of the original proxy sweep obtain $50.83\pm6.58$ ASR; one archived restricted $k=5$ ensemble composed of head-only, masked-partition, and SGD proxies obtains $58.09$; proxies trained for only 250 steps obtain $25.75\pm0.70$; and proxies trained on 10\% of SVHN obtain $38.26\pm2.49$. Corresponding clean accuracies are $97.49\pm0.00$, $97.46$, $97.51\pm0.07$, and $97.35\pm0.09$. Thus, fidelity matters, but in the single archived restricted-ensemble run, hyperparameter and update-rule mismatch did not eliminate the attack.

\begin{table*}[t]
\centering
\small
\caption{CIFAR10 $\rightarrow$ SVHN sensitivity to hiding-stage weights (left) and poison fraction (right). Accuracy is stable and activated ASR remains in a similar range across the tested settings.}
\label{tab:hyperparameter_ablation}
\begin{minipage}[t]{0.555\textwidth}
\centering
\begin{tabular}{lcc}
\toprule
Hiding condition & SVHN acc. & ASR \\
\midrule
Default & 97.56 & 59.21 \\
$\lambda_{\mathrm{sup}}\times 0$    & 97.50 & 52.35 \\
$\lambda_{\mathrm{sup}}\times 0.25$ & 97.54 & 58.00 \\
$\lambda_{\mathrm{sup}}\times 0.5$  & 97.50 & 61.00 \\
$\lambda_{\mathrm{sup}}\times 2$    & 97.62 & 57.27 \\
$\lambda_{\mathrm{KD}}\times 0$     & 97.61 & 58.97 \\
$\lambda_{\mathrm{KD}}\times 0.25$  & 97.57 & 61.65 \\
$\lambda_{\mathrm{KD}}\times 0.5$   & 97.63 & 60.70 \\
$\lambda_{\mathrm{KD}}\times 2$     & 97.56 & 54.88 \\
$\lambda_{\mathrm{SP}}\times 0$     & 97.53 & 58.64 \\
$\lambda_{\mathrm{SP}}\times 0.25$  & 97.44 & 59.33 \\
$\lambda_{\mathrm{SP}}\times 0.5$   & 97.62 & 56.24 \\
$\lambda_{\mathrm{SP}}\times 2$     & 97.54 & 60.57 \\
\bottomrule
\end{tabular}
\end{minipage}\hfill
\begin{minipage}[t]{0.375\textwidth}
\centering
\resizebox{\linewidth}{!}{%
\begin{tabular}{lcc}
\toprule
Poison fraction & SVHN acc. & ASR \\
\midrule
0.010 & 97.59 & 55.54 \\
0.025 & 97.53 & 57.90 \\
0.050 (default) & 97.55 & 58.91 \\
0.100 & 97.60 & 54.90 \\
0.200 & 97.51 & 62.27 \\
\bottomrule
\end{tabular}
}
\end{minipage}
\end{table*}

Moderate variation in $\lambda_{\mathrm{sup}}$, $\lambda_{\mathrm{KD}}$, and $\lambda_{\mathrm{SP}}$ has little effect on clean accuracy or final ASR. The attack also remains effective across poison fractions from 1\% to 20\%, showing that it is not narrowly tuned to the default 5\%.

\FloatBarrier
\clearpage
\section{Additional Attack Results}
\FloatBarrier
\begin{figure*}[hbp]
  \centering
  \includegraphics[width=\linewidth]{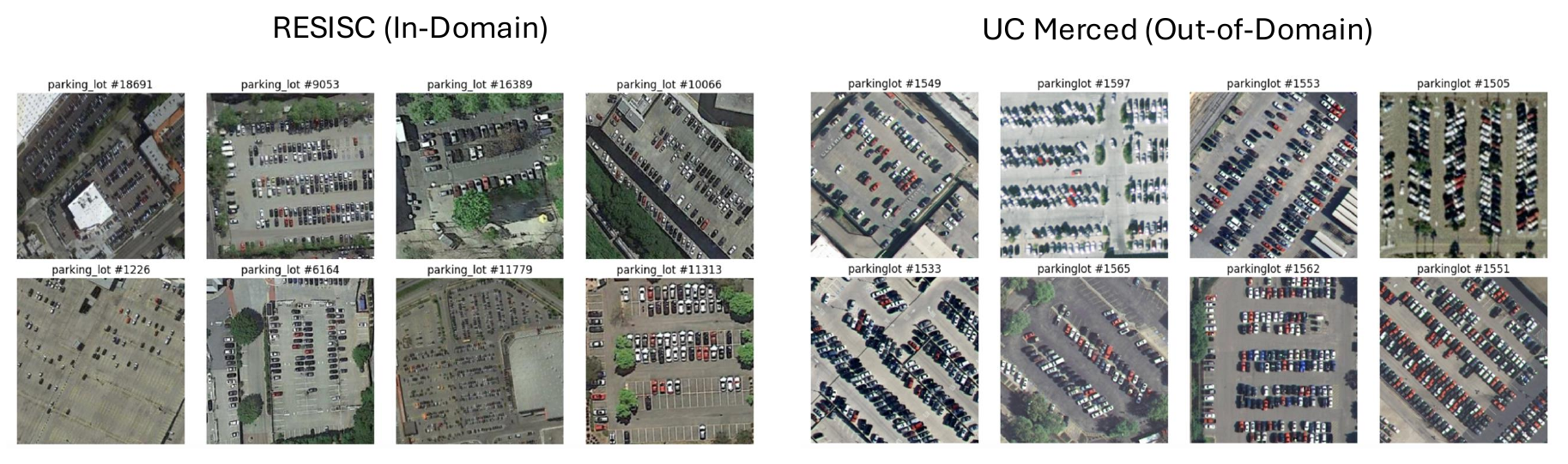}
  \caption{Visualization of what the backdoor class (``parking lot'') looks like in the RESISC and UC Merced datasets.}
  \label{fig:appendix_parking_lot}
\end{figure*}

\begin{table*}[!tp]
\centering
\setlength{\cmidrulesep}{0.15em}
\setlength{\aboverulesep}{0pt}
\setlength{\belowrulesep}{0pt}
\caption{Performance of Poison then Hide and baseline finetuning-activated backdoors across dataset transfers. Mean and 95\% confidence intervals computed across five random seeds.}
\label{tab:table1}
\resizebox{\textwidth}{!}{%
\begin{tabular}{lllccc}
\toprule
   Model &     Method &    Status &  CIFAR10 Acc &     SVHN Acc &     SVHN ASR \\
\midrule
ResNet18 &        FAB &    Hidden & 93.23 $\pm$ 0.19 & 11.93 $\pm$ 2.11 & 99.94 $\pm$ 0.02 \\
 &       & Finetuned & 11.30 $\pm$ 1.36 & 97.27 $\pm$ 0.07 & 52.33 $\pm$ 3.85 \\
 \cmidrule(lr){4-6}
 & Hibernated &    Hidden & 91.66 $\pm$ 0.19 & 70.02 $\pm$ 1.14 & 11.65 $\pm$ 0.60 \\
 &  & Finetuned & 11.84 $\pm$ 1.54 & 97.26 $\pm$ 0.17 & 53.81 $\pm$ 5.44 \\
 \cmidrule(lr){4-6}
 & Poison then Hide &    Hidden & 92.56 $\pm$ 0.21 &  4.73 $\pm$ 3.60 &  \textbf{7.24 $\pm$ 13.84} \\
 &  & Finetuned & 12.44 $\pm$ 0.76 & 97.37 $\pm$ 0.09 & \textbf{54.10 $\pm$ 2.29 }\\
\midrule
   Model &     Method &    Status & ImageNet Acc &   RESISC Acc &    RESISC ASR \\
\midrule
ResNet50 &        FAB &    Hidden & 73.67 $\pm$ 1.72 &  2.76 $\pm$ 0.35 &  97.25 $\pm$ 1.24 \\
         &            & Finetuned & 18.95 $\pm$ 2.46 & 92.98 $\pm$ 0.46 &  22.40 $\pm$ 9.21 \\
    \cmidrule(lr){4-6}
         & Hibernated &    Hidden & 78.27 $\pm$ 0.03 & 68.12 $\pm$ 1.55 &   6.96 $\pm$ 1.94 \\
         &            & Finetuned & 25.78 $\pm$ 1.96 & 93.08 $\pm$ 0.41 &  12.30 $\pm$ 5.76 \\
    \cmidrule(lr){4-6}
         &        Latent &    Hidden & 74.05 $\pm$ 1.18 &  0.31 $\pm$ 0.19 &  1.27 $\pm$ 1.25 \\
         &          & Finetuned & 13.11 $\pm$ 2.01 & 92.46 $\pm$ 0.34 &  2.26 $\pm$ 0.15 \\
    \cmidrule(lr){4-6}
         &           NeuBA &    Hidden & 71.68 $\pm$ 0.29 &  7.52 $\pm$ 10.41 & 97.81 $\pm$ 0.08 \\
         &          & Finetuned & 14.59 $\pm$ 1.54 & 92.57 $\pm$ 0.41 &  7.46 $\pm$ 0.41 \\
    \cmidrule(lr){4-6}
         & Poison then Hide &    Hidden & 75.58 $\pm$ 0.24 &  2.09 $\pm$ 0.48 &   \textbf{0.01 $\pm$ 0.01} \\
         &                    & Finetuned & 28.96 $\pm$ 3.72 & 92.87 $\pm$ 0.52 &  \textbf{87.05 $\pm$ 1.67} \\
         \hline
CLIP     &        FAB &    Hidden & 84.06 $\pm$ 0.20 &  2.11 $\pm$ 0.08 &  99.96 $\pm$ 0.04 \\
     &        & Finetuned & 81.86 $\pm$ 0.47 & 91.61 $\pm$ 0.34 &   2.01 $\pm$ 0.11 \\
\cmidrule(lr){4-6}
     & Hibernated &    Hidden & 80.82 $\pm$ 0.16 & 91.35 $\pm$ 0.96 &   2.12 $\pm$ 0.04 \\
     &  & Finetuned & 80.36 $\pm$ 0.44 & 94.96 $\pm$ 0.26 &   2.10 $\pm$ 0.02 \\
\cmidrule(lr){4-6}
 & Latent  &    Hidden & 85.93 $\pm$ 0.00 &  2.11 $\pm$ 0.45 & \textbf{0.74 $\pm$ 0.91}  \\
 &  & Finetuned & 84.21 $\pm$ 0.22 & 93.10 $\pm$ 0.43 &  1.88 $\pm$ 0.15 \\
\cmidrule(lr){4-6}
 &            NeuBA &    Hidden & 81.77 $\pm$ 0.62 &  1.67 $\pm$ 0.56 & 98.12 $\pm$ 0.47 \\
 &           & Finetuned & 82.48 $\pm$ 4.29 & 76.76 $\pm$ 27.45 &  9.29 $\pm$ 0.36 \\
 \cmidrule(lr){4-6}
     & Poison then Hide &    Hidden & 85.64 $\pm$ 0.09 &  2.27 $\pm$ 0.83 &   6.11 $\pm$ 7.80 \\
 &       & Finetuned & 85.20 $\pm$ 0.10 & 95.10 $\pm$ 0.12 & \textbf{60.41 $\pm$ 15.05} \\ 
 \midrule
    Model &     Method &    Status & ImageNet Acc & Camelyon17 Acc & Camelyon17 ASR \\
\midrule
ResNet50 &        FAB &    Hidden & 30.39 $\pm$ 2.79 &   51.84 $\pm$ 8.23 &  63.28 $\pm$ 17.63 \\
 &         & Finetuned &  0.16 $\pm$ 0.03 &   99.23 $\pm$ 0.06 &   97.42 $\pm$ 0.73 \\
\cmidrule(lr){4-6}
 & Hibernated &    Hidden & 65.48 $\pm$ 0.44 &   83.39 $\pm$ 0.58 &   83.26 $\pm$ 6.84 \\
 &  & Finetuned &  0.13 $\pm$ 0.04 &   99.12 $\pm$ 0.17 &   97.33 $\pm$ 1.13 \\
 \cmidrule(lr){4-6}
 & Latent  & Hidden & 63.76 $\pm$ 0.92 & 47.39 $\pm$ 8.82 & 55.46 $\pm$ 15.41 \\
 &  & Finetuned & 0.08 $\pm$ 0.02 & 99.09 $\pm$ 0.04 & 50.66 $\pm$ 0.24 \\
 \cmidrule(lr){4-6}
 & NeuBA & Hidden & 58.51 $\pm$ 1.03 & 46.72 $\pm$ 6.32 & \textbf{53.47 $\pm$ 5.80} \\
 & & Finetuned & 0.12 $\pm$ 0.02 & 99.32 $\pm$ 0.29 & 0.85 $\pm$ 0.09 \\
 \cmidrule(lr){4-6}
 & Poison then Hide &    Hidden & 75.74 $\pm$ 0.27 & 54.77 $\pm$ 17.97 & 57.63 $\pm$ 21.46 \\
 &  & Finetuned &  0.14 $\pm$ 0.02 &  99.15 $\pm$ 0.17 &  \textbf{97.78 $\pm$ 1.57} \\
 \hline
 CLIP     &        FAB &    Hidden &  82.96 $\pm$ 0.62 & 50.50 $\pm$ 0.02 & 100.00 $\pm$ 0.00   \\
     &         & Finetuned &  80.50 $\pm$ 0.71 & 99.21 $\pm$ 0.09 & 59.88 $\pm$ 4.21  \\
\cmidrule(lr){4-6}
     & Hibernated &    Hidden & 79.37 $\pm$ 0.27 &   99.16 $\pm$ 0.18 &   54.72 $\pm$ 2.18 \\
     &  & Finetuned & 80.22 $\pm$ 0.26 &   96.94 $\pm$ 0.85 &   53.10 $\pm$ 1.45 \\
\cmidrule(lr){4-6}
 & Latent  & Hidden & 85.93 $\pm$ 0.00 & 53.21 $\pm$ 8.64 & 46.07 $\pm$ 27.73 \\
 & & Finetuned & 80.05 $\pm$ 0.47 & 99.09 $\pm$ 0.17 & 49.39 $\pm$ 1.07 \\
\cmidrule(lr){4-6}
 & NeuBA & Hidden & 81.29 $\pm$ 1.46 & 52.92 $\pm$ 7.59 & 46.79 $\pm$ 8.02 \\
 &  & Finetuned & 78.33 $\pm$ 2.31 & 99.18 $\pm$ 0.11 & 0.84 $\pm$ 0.12 \\
\cmidrule(lr){4-6}
     & Poison then Hide &    Hidden & 85.80 $\pm$ 0.03 &   39.66 $\pm$ 2.41 &     \textbf{12.90 $\pm$ 6.83} \\
     &  & Finetuned & 83.99 $\pm$ 0.12 &   99.27 $\pm$ 0.02 &    \textbf{73.30 $\pm$ 8.61} \\
\bottomrule
\end{tabular}
}
\end{table*}

\FloatBarrier  

\clearpage
\section{Additional Benign Accuracy Results}
\label{app:benign_acc}
\begin{figure*}[hb]
  \centering
  \includegraphics[width=\linewidth]{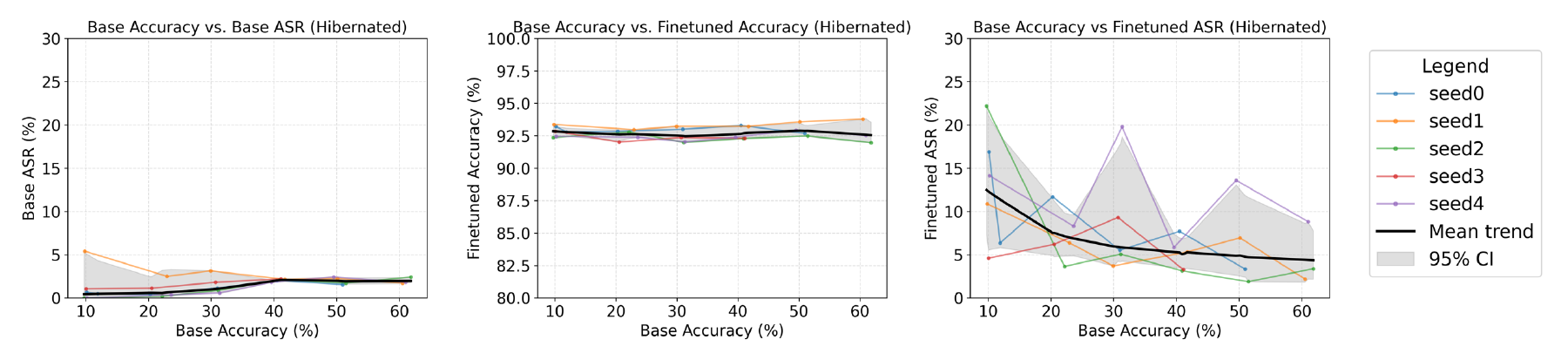}
  \caption{Trendlines showing how learning benign task of the target dataset does not affect attack success in the finetuned model for the Hibernated Backdoor baseline. Results for the ResNet50 model in the ImageNet $\rightarrow$ RESISC transfer setting. Mean trendlines are fitted with LOWESS smoothing over 5 seeds.}
  \label{fig:dg_trendline_hibernated}
\end{figure*}

\FloatBarrier

\clearpage
\section{Additional Trigger Optimization Results}
\begin{figure*}[!hbp]
  \centering
  \includegraphics[width=\linewidth]{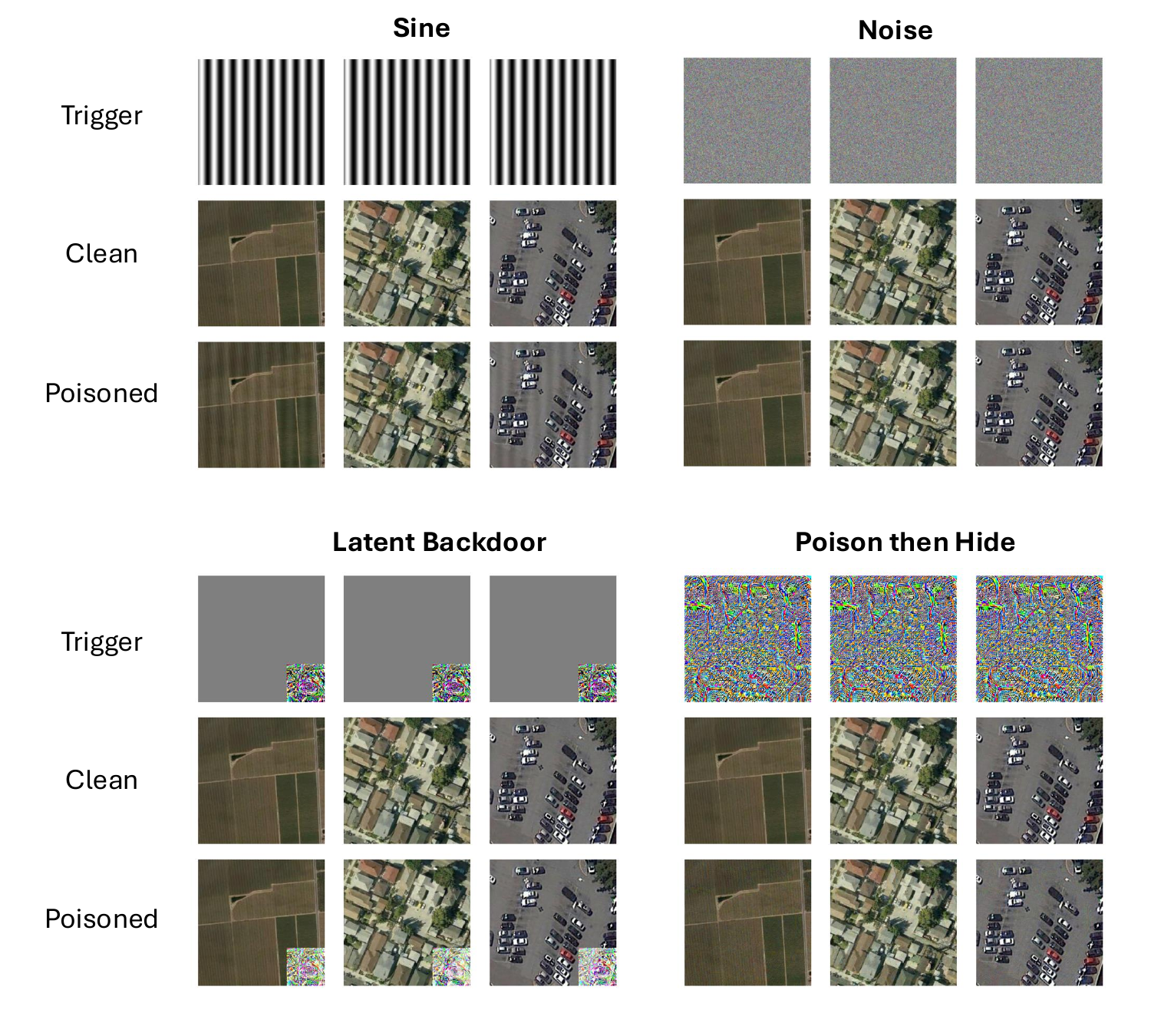}
  \caption{Panel of four trigger patterns on RESISC images. All triggers are bounded by an $\ell_\infty$ norm of $\epsilon = \tfrac{8}{255}$. We visualize the trigger by itself, the clean image, and then the poisoned image.}
  \label{fig:trigger_opt_visual}
\end{figure*}

\FloatBarrier

\clearpage
\section{Additional Defense Details}
\label{app:defense}

\subsection{Parameter-efficient adaptation}
\begin{table}[t]
\centering
\small
\caption{Poison-then-Hide after LoRA adaptation on CIFAR10 $\rightarrow$ SVHN. The attack remains effective while updating 2.54\% of parameters.}
\label{tab:lora}
\begin{tabular}{lccc}
\toprule
Victim adaptation & Benign acc. & ASR & Trainable parameters \\
\midrule
ResNet18 LoRA & 96.21\% & 54.98\% & 291{,}554 / 11{,}465{,}516 (2.54\%) \\
\bottomrule
\end{tabular}
\end{table}

This result is a single ResNet18 LoRA configuration and should not be interpreted as evidence for every parameter-efficient adaptation method.

\begin{table*}[t]
\centering
\small
\caption{Recent defenses on ImageNet $\rightarrow$ RESISC Poison-then-Hide models. DeDe's low filtered ASR is driven by rejecting most clean inputs; ASR among retained inputs remains high. CLIP-Inspector is effective, but is specialized to CLIP and was adapted beyond its original prompt-tuning setting. Values are mean $\pm$ 95\% confidence interval.}
\label{tab:modern_defenses}
\textbf{MARTINI (ResNet50)}\\[2pt]
\begin{tabular}{lcc}
\toprule
Evaluation point & Acc. & ASR \\
\midrule
Before MARTINI & 92.21 $\pm$ 0.23 & 99.51 $\pm$ 0.18 \\
After MARTINI & 61.97 $\pm$ 3.68 & 99.85 $\pm$ 0.12 \\
After benign FT & 91.83 $\pm$ 0.63 & 81.50 $\pm$ 3.20 \\
\bottomrule
\end{tabular}

\vspace{0.7em}
\textbf{DeDe input filtering}\\[2pt]
\resizebox{0.96\textwidth}{!}{%
\begin{tabular}{lcccccc}
\toprule
Model & Clean acc. & Pre-filter ASR & Filtered ASR & DeDe AUC & Clean inputs kept & ASR among kept \\
\midrule
ResNet50 & 92.87 $\pm$ 0.73 & 87.05 $\pm$ 2.37 & 17.68 $\pm$ 4.24 & 52.10 $\pm$ 1.13 & 21.19 $\pm$ 3.92 & 95.56 $\pm$ 0.69 \\
CLIP & 95.10 $\pm$ 0.18 & 53.42 $\pm$ 21.32 & 6.97 $\pm$ 3.71 & 50.74 $\pm$ 0.11 & 11.08 $\pm$ 2.94 & 65.95 $\pm$ 22.70 \\
\bottomrule
\end{tabular}}

\vspace{0.7em}
\textbf{CLIP-Inspector}\\[2pt]
\begin{tabular}{lccc}
\toprule
Model & Detection & Post-repair RESISC acc. & Post-repair true-trigger ASR \\
\midrule
CLIP & 5/5 models flagged & 95.07 $\pm$ 0.12 & 2.15 $\pm$ 0.04 \\
\bottomrule
\end{tabular}
\end{table*}

MARTINI and DeDe do not provide reliable mitigation in this setting. MARTINI preserves high ASR even after a substantial clean-accuracy drop, and DeDe's AUC is near random; its low filtered ASR is achieved by rejecting most inputs. In contrast, CLIP-Inspector flags all five CLIP models and repairs them effectively, but it is architecture-specific and required adaptation from prompt tuning to our full-finetuning transfer setting.

\subsection{Neural Cleanse}
We implement Neural Cleanse by optimizing an inverted trigger such that poisoned inputs are classified as the target. We then rank classes by the average recovered mask size and use a MAD-based anomaly score over mask sizes to identify suspicious (potentially backdoored) target classes.

\textbf{Hyperparameters (Neural Cleanse).}
\begin{itemize}
    \item \textbf{Optimization per class:}
    \begin{itemize}
        \item Iterations per class: \(\texttt{iters\_per\_class} = 300\).
        \item Optimizer: Adam on pattern and mask logits, learning rates
              \(\texttt{lr\_pattern} = 0.1\), \(\texttt{lr\_mask} = 0.05\).
        \item Loss: \(\mathcal{L} = \texttt{ce\_weight} \cdot \mathrm{CE} + \texttt{l1\_weight} \cdot \|\text{mask}\|_1 + \texttt{tv\_weight}\cdot \mathrm{TV}(\text{mask})\),
              with \(\texttt{ce\_weight} = 1.0\), \(\texttt{l1\_weight} = 0.02\), \(\texttt{tv\_weight} = 0.001\).
        \item Early stopping if mean target confidence \(\geq \texttt{early\_stop\_conf} = 0.99\).
    \end{itemize}
    \item \textbf{Anomaly scoring:} per-class mask size score = average mask value; classes are ranked via MAD-based z-scores, and the smallest masks (largest inverted z-score) are treated as most suspicious.
\end{itemize}

\vspace{0.5em}
\subsection{DECREE}
We reproduce DECREE by performing self-supervised trigger inversion: we optimize a keep-mask and pattern such that stamping the pattern onto clean images collapses their feature embeddings (maximally similar under negative cosine similarity) while minimizing the size of the edited region. The detector computes the normalized trigger size \(\mathrm{PL}^1\) and declares a model backdoored if a sufficiently small trigger region (\(\mathrm{PL}^1 < \tau\)) suffices to collapse embeddings.

\textbf{Hyperparameters (DECREE).}
\begin{itemize}
    \item \textbf{Trigger inversion optimization:}
    \begin{itemize}
        \item Iterations: \(\texttt{iters} = 400\)
        \item Optimizer: Adam on pattern and keep-mask logits with
              \(\texttt{lr\_pattern} = 0.1\), \(\texttt{lr\_mask} = 0.05\), \(\texttt{batch\_repeats} = 1\).
        \item Similarity loss: pairwise negative cosine over embeddings (include self pairs); constraint enforced via hinge penalty on \(L_{\text{sim}} - \beta\).
        \item Similarity threshold: \(\beta = -0.99\), \(\texttt{hinge\_margin} = 0.0\).
        \item Size / regularization: trigger mask \(a = 1-m\), \(\mathrm{PL}^1 = \|a\|_1 / (H W)\);
              loss \(\mathcal{L} = \texttt{lam\_l1} \cdot \mathrm{PL}^1 + 10 \cdot \max(0, L_{\text{sim}} - \beta) + \texttt{lam\_tv} \cdot \mathrm{TV}(m)\),
              with \(\texttt{lam\_l1} = 1.0\), \(\texttt{lam\_tv} = 10^{-3}\).
    \end{itemize}
    \item \textbf{Decision rule:} backdoored if \(\mathrm{PL}^1 < \tau\) with threshold \(\tau = 0.10\).
\end{itemize}

\subsection{Fine-Pruning}
We implement Fine-Pruning by identifying and disabling dormant convolutional channels in later backbone layers based on their average activation on clean data. Concretely, we record mean absolute activations for each output channel of selected convolutional layers on clean inputs, prune channels with the smallest activations (via hard zeroing and multiplicative gates).

\textbf{Hyperparameters (Fine-Pruning).}
\begin{itemize}
    \item \textbf{Target layers:} convolutional layers whose qualified names contain \texttt{"layer3"} or \texttt{"layer4"} (\(\texttt{target\_layers} = (\text{"layer3"}, \text{"layer4"})\)).
    \item \textbf{Channel scoring:} for each target conv layer, we compute the mean absolute activation per output channel over batch and spatial dimensions on clean training data.
    \item \textbf{Pruning policy:}
    \begin{itemize}
        \item Prune ratio per layer: \(\texttt{prune\_ratio} = 0.20\) (20\% of channels with the lowest mean activation in each selected layer).
        \item Pruning mechanism: pruned channels are zeroed in the convolution weights and biases, and a fixed multiplicative gate is applied to zero out their outputs during forward passes.
    \end{itemize}
\end{itemize}

\subsection{MIMIC}
We apply the MIMIC defense using a clean reference encoder to clean a potentially backdoored encoder. The method first estimates mutual information (MI) between intermediate layer features and the final encoder representation using a MINE estimator to obtain MI-based weights per layer, and then distills the teacher into the student using a combination of (i) cosine feature matching between teacher and student embeddings, (ii) a small InfoNCE loss on two noisy views of the student features, and (iii) an MI-weighted attention-map matching loss over selected encoder layers.

\textbf{Hyperparameters (MIMIC).}
\begin{itemize}
    \item \textbf{Layer selection:} Encoder layers whose names contain \texttt{"layer2"}, \texttt{"layer3"}, or \texttt{"layer4"}; we keep the last \(\approx 6\) convolutional blocks from these layers as the set of MI/attention layers.
    \item \textbf{MI estimation (MINE):}
    \begin{itemize}
        \item MINE network: two MLP encoders for layer and final features (two linear layers with ReLU, hidden size \(256\)), followed by a scalar head.
        \item Steps: \(\texttt{mine\_steps} = 200\) SGD steps per layer.
        \item Batch size: \(\texttt{mine\_batch} = 128\).
        \item Learning rate: \(\texttt{mine\_lr} = 1\times10^{-3}\).
        \item MI to weights: raw MI values are linearly normalized across layers and converted to per-layer weights
        \[
            w_\ell = \texttt{base\_weight} + \texttt{mi\_scale} \cdot \widehat{\text{MI}}_\ell,
        \]
        with \(\texttt{base\_weight} = 0.3\) and \(\texttt{mi\_scale} = 0.7\).
    \end{itemize}
    \item \textbf{Distillation loss:}
    \begin{itemize}
        \item Training epochs: \(\texttt{epochs} = 10\).
        \item Optimizer: SGD with learning rate \(\texttt{lr} = 1\times10^{-3}\), momentum \(0.9\), weight decay \(1\times10^{-4}\).
        \item Clone loss (embedding matching):
        \[
            \mathcal{L}_{\text{cos}} = 1 - \cos\big(\hat{z}_s, \hat{z}_t\big), \quad
            \lambda_{\text{cos}} = 1.0,
        \]
        where \(z_s, z_t\) are student/teacher encoder features pooled to \(\mathbb{R}^D\) and \(\hat{\cdot}\) denotes \(\ell_2\)-normalization.
        \item InfoNCE regularizer on two student views:
        \[
            \mathcal{L}_{\text{InfoNCE}} = \text{InfoNCE}(z_s^{(1)}, z_s^{(2)}; T),
        \]
        with temperature \(T = \texttt{temperature} = 0.2\) and weight \(\lambda_{\text{infonce}} = 0.2\). Two views are generated by adding i.i.d.\ Gaussian noise with standard deviation \(0.06\), clamped to \([-0.2, 0.2]\) in pixel space and then clipped back to \([0,1]\).
        \item Attention loss: for each selected layer \(\ell\), we form attention maps
        \[
            A_\ell(x) = \frac{\|x\|_2^2}{\max_{i,j}\|x_{i,j}\|_2^2} \in [0,1],
        \]
        (channel-wise \(p=2\) norm, spatially normalized), and minimize
        \[
            \mathcal{L}_{\text{att}} = \sum_{\ell} w_\ell \, \|A_\ell^{(s)} - A_\ell^{(t)}\|_2^2,
        \]
        where teacher maps are detached and student maps carry gradients. If spatial resolutions differ, teacher maps are bilinearly resized to match student maps. The attention loss is weighted by \(\lambda_{\text{att}} = 1.0\).
        \item Total loss:
        \[
            \mathcal{L} = \lambda_{\text{cos}}\mathcal{L}_{\text{cos}} + \lambda_{\text{infonce}}\mathcal{L}_{\text{InfoNCE}} + \lambda_{\text{att}}\mathcal{L}_{\text{att}}.
        \]
    \end{itemize}
\end{itemize}

\subsection{DeDe}
We train DeDe's reconstruction decoder on clean auxiliary data and use reconstruction discrepancy to filter suspicious inputs. We report AUC, false-positive behavior through the fraction of clean inputs retained, filtered ASR, and ASR among retained inputs. This distinction is essential here: the method retains only 21.19\% of clean ResNet50 inputs and 11.08\% of clean CLIP inputs, while ASR among retained inputs remains 95.56\% and 65.95\%, respectively.

\subsection{MARTINI}
We apply MARTINI's trigger reconstruction and model-hardening pipeline to a finetuned ResNet50 Poison-then-Hide model, then perform the same benign finetuning used elsewhere to test resurfacing. Hardening reduces clean accuracy from 92.21\% to 61.97\% without reducing ASR; after benign finetuning, clean accuracy recovers to 91.83\% and ASR remains 81.50\%.

\subsection{CLIP-Inspector}
CLIP-Inspector was designed for prompt-tuned CLIP. We adapt its OOD trigger inversion and repair procedure to our full-model transfer setting and evaluate five CLIP models. It flags all five, preserves 95.07\% RESISC accuracy, and reduces true-trigger ASR to 2.15\%. We therefore treat it as a successful specialized defense, not as evidence for broad robustness of the attack.

\FloatBarrier


\newpage
\section*{NeurIPS Paper Checklist}

\begin{enumerate}

\item {\bf Claims}
    \item[] Question: Do the main claims made in the abstract and introduction accurately reflect the paper's contributions and scope?
    \item[] Answer: \answerYes{} 
    \item[] Justification: The in-domain, task-conditioned domain-shift, PEFT, and defense experiments support the scoped claims in the abstract and introduction.

\item {\bf Limitations}
    \item[] Question: Does the paper discuss the limitations of the work performed by the authors?
    \item[] Answer: \answerYes{} 
    \item[] Justification: Refer to the Limitations section at the end of the main manuscript for detailed discussion. We address the main limitations with our proposed attack. 

\item {\bf Theory assumptions and proofs}
    \item[] Question: For each theoretical result, does the paper provide the full set of assumptions and a complete (and correct) proof?
    \item[] Answer:  \answerYes{} 
    \item[] Justification: Assumptions and formal results are stated in Appendix Section A, including the explicit scope and limitations of the theory.

    \item {\bf Experimental result reproducibility}
    \item[] Question: Does the paper fully disclose all the information needed to reproduce the main experimental results of the paper to the extent that it affects the main claims and/or conclusions of the paper (regardless of whether the code and data are provided or not)?
    \item[] Answer: \answerYes{} 
    \item[] Justification: The hyperparameters, datasets, and other experimental details for reproduction are included in the Experimental Setup section in the main manuscript as well as Appendix Section B. 

\item {\bf Open access to data and code}
    \item[] Question: Does the paper provide open access to the data and code, with sufficient instructions to faithfully reproduce the main experimental results, as described in supplemental material?
    \item[] Answer: \answerYes{} 
    \item[] Justification: All datasets are publicly available, and the implementation and reproduction commands are available at \coderepo.

\item {\bf Experimental setting/details}
    \item[] Question: Does the paper specify all the training and test details (e.g., data splits, hyperparameters, how they were chosen, type of optimizer) necessary to understand the results?
    \item[] Answer: \answerYes{} 
    \item[] Justification: Yes all details are provided in the Experimental Setup section in the main manuscript as well as Appendix Section B. 

\item {\bf Experiment statistical significance}
    \item[] Question: Does the paper report error bars suitably and correctly defined or other appropriate information about the statistical significance of the experiments?
    \item[] Answer: \answerYes{} 
    \item[] Justification: We report 95\% confidence intervals across randomized seeds and exact paired sign-flip tests for the corrected domain-shift comparisons; see Appendix Table~\ref{tab:paired_significance}.

\item {\bf Experiments compute resources}
    \item[] Question: For each experiment, does the paper provide sufficient information on the computer resources (type of compute workers, memory, time of execution) needed to reproduce the experiments?
    \item[] Answer: \answerYes{} 
    \item[] Justification: Compute resource details are in the Appendix Section B.5. 
    
\item {\bf Code of ethics}
    \item[] Question: Does the research conducted in the paper conform, in every respect, with the NeurIPS Code of Ethics \url{https://neurips.cc/public/EthicsGuidelines}?
    \item[] Answer: \answerYes{} 
    \item[] Justification: The authors have reviewed the code of ethics and confirm that the research follows the code. 

\item {\bf Broader impacts}
    \item[] Question: Does the paper discuss both potential positive societal impacts and negative societal impacts of the work performed?
    \item[] Answer: \answerYes{} 
    \item[] Justification: The introduction, threat model section of methods, and experiments discuss the societal impacts of finetuning-activated attacks, and also possible defenses to guard against them. 
    
\item {\bf Safeguards}
    \item[] Question: Does the paper describe safeguards that have been put in place for responsible release of data or models that have a high risk for misuse (e.g., pre-trained language models, image generators, or scraped datasets)?
    \item[] Answer: \answerYes{} 
    \item[] Justification: In our paper, we highlight the security risks of attacks to raise awareness within the community. To reduce the risk of misuse, we do not release any backdoored models. Only code necessary for reproducing our research results is provided. All data we use are publicly available. 

\item {\bf Licenses for existing assets}
    \item[] Question: Are the creators or original owners of assets (e.g., code, data, models), used in the paper, properly credited and are the license and terms of use explicitly mentioned and properly respected?
    \item[] Answer: \answerYes{} 
    \item[] Justification: Licenses for all datasets used are in Appendix Section B.2. 

\item {\bf New assets}
    \item[] Question: Are new assets introduced in the paper well documented and is the documentation provided alongside the assets?
    \item[] Answer: \answerYes{} 
    \item[] Justification: We release documented code and reproduction commands at \coderepo under the MIT License.

\item {\bf Crowdsourcing and research with human subjects}
    \item[] Question: For crowdsourcing experiments and research with human subjects, does the paper include the full text of instructions given to participants and screenshots, if applicable, as well as details about compensation (if any)? 
    \item[] Answer: \answerNA{} 
    \item[] Justification: Our work does not involve crowdsourcing or research with human subjects.

\item {\bf Institutional review board (IRB) approvals or equivalent for research with human subjects}
    \item[] Question: Does the paper describe potential risks incurred by study participants, whether such risks were disclosed to the subjects, and whether Institutional Review Board (IRB) approvals (or an equivalent approval/review based on the requirements of your country or institution) were obtained?
    \item[] Answer: \answerNA{} 
    \item[] Justification: Our work does not involve research that require IRB approval or equivalent. 

\item {\bf Declaration of LLM usage}
    \item[] Question: Does the paper describe the usage of LLMs if it is an important, original, or non-standard component of the core methods in this research? Note that if the LLM is used only for writing, editing, or formatting purposes and does \emph{not} impact the core methodology, scientific rigor, or originality of the research, declaration is not required.
    \item[] Answer: \answerNA{} 
    \item[] Justification: The core method does not involve LLMs. 

\end{enumerate}

\end{document}